\documentclass{article} 
\usepackage{iclr2027_conference,times}

\usepackage{amsmath,amsfonts,bm}

\def\eqref#1{equation~\ref{#1}}

\def\1{\bm{1}}

\DeclareMathAlphabet{\mathsfit}{\encodingdefault}{\sfdefault}{m}{sl}
\SetMathAlphabet{\mathsfit}{bold}{\encodingdefault}{\sfdefault}{bx}{n}

\usepackage{hyperref}
\usepackage{url}
\usepackage{multirow}
\usepackage{graphicx}
\usepackage[table]{xcolor}
\usepackage{duckuments}
\usepackage{paralist}
\usepackage{bigdelim}   
\usepackage{caption}
\usepackage{subfigure}
\usepackage{cleveref}
\usepackage{booktabs}
\usepackage{wrapfig}

\crefname{section}{Sec.}{Secs.}
\Crefname{section}{Sec.}{Secs.}
\crefname{subsection}{Sec.}{Secs.}
\Crefname{subsection}{Sec.}{Secs.}
\crefname{table}{Tab.}{Tabs.}
\Crefname{table}{Tab.}{Tabs.}
\crefname{figure}{Fig.}{Figs.}
\Crefname{figure}{Fig.}{Figs.}
\crefname{equation}{Eq.}{Eqs.}
\Crefname{equation}{Eq.}{Eqs.}
\crefname{appendix}{App.}{Apps.}
\Crefname{appendix}{App.}{Apps.}

\title{Natural Image Autoencoder-Based fMRI Representations for Trait and State Prediction}

\newcommand{\meanstd}[2]{$#1_{\pm #2}$}

\newcommand{\oursl}{FReD-Trait}
\newcommand{\ourst}{FReD-State}
\newcommand{\ours}{FReD}

\author{
Juhyeon Park$^{1,3}$\thanks{Work done during an internship at Microsoft Research.}, Yeonwoo Kim$^{1}$, Peter Yongho Kim$^{2,3}$\footnotemark[1], Yansen Wang$^{3}$, \\
\textbf{Mingqing Xiao$^{3}$, Dongqi Han$^{3}$, Dongsheng Li$^{3}$,
Taesup Moon$^{1,2,4}$\thanks{Corresponding author.}} \\
$^{1}$IPAI, Seoul National University,
$^{2}$ECE, Seoul National University, \\
$^{3}$Microsoft Research,
$^{4}$ASRI / INMC / AIIS, Seoul National University\\
\texttt{\{parkjh9229, yeonwoo4, peterkim98, tsmoon\}@snu.ac.kr},\\ 
\texttt{\{yansenwang, mingqingxiao, dongqihan, dongsli\}@microsoft.com}
}

\iclrfinalcopy 
\begin{document}

\maketitle

\begin{abstract}
Foundation models pre-trained on large-scale fMRI datasets have shown strong downstream performance, but at substantial data and computation cost. To investigate how much fMRI-specific pre-training is actually needed for such performance, we introduce {\ours}, which derives fMRI representations from a frozen Deep Compression AutoEncoder (DCAE) pre-trained exclusively on natural images and pairs them with a task specific readout. For trait prediction, {\ours} summarizes frame-wise representations by their temporal mean and log-standard deviation and applies linear probing, with late fusion across two normalization schemes. For state prediction, it represents each frame as a single token and models temporal dependencies with a shallow Transformer. Across four resting-state datasets spanning six trait-prediction targets, linear probes on frozen DCAE features generally outperform those on fMRI foundation model representations and remain competitive with fully fine-tuned fMRI foundation models. On three task-fMRI state-prediction tasks, a temporal readout on DCAE features performs comparably to the strongest foundation models evaluated. A Gaussian injection analysis further shows that localized signal changes are recovered more accurately from the frozen DCAE features than from the evaluated foundation-model representations. Together, these results show that strong performance on current fMRI benchmarks is possible without fMRI-specific representation pre-training, making frozen natural-image features as a useful baseline for assessing its added value.
\end{abstract}
\section{Introduction}
\label{sec:intro}

Deep learning-based approaches to fMRI signal processing have been
widely explored.
More recently, foundation models pre-trained on large-scale fMRI datasets have demonstrated strong performance across a broad range of downstream tasks~\citep{swift, brainjepa, braindit, cortexmae, neurostorm, omnifmri, brainmass}.
Developing such models, however, typically relies on large-scale neuroimaging datasets that are costly to collect and preprocess, while the pre-training itself may also require substantial computational resources.

One alternative is to reuse representations learned outside the fMRI domain. TABLeT~\citep{tablet}, for example, showed that Deep Compression AutoEncoder (DCAE)~\citep{dcae}, pre-trained exclusively on natural images, can be applied to fMRI data and used for modeling long-range temporal dynamics, suggesting that an encoder trained without any fMRI data may already provide a useful starting point. How far such a frozen representation can go, however, remains unclear. TABLeT considers only volumetric fMRI and represents each frame as multiple tokens so that its Transformer \citep{transformer} must model spatial and temporal relationships jointly. It also applies the same tokenization and temporal modeling to every downstream task; a single uniform recipe is appealing, but whether it is adequate has not been tested, since trait and state prediction differ in whether the target varies within a subject.
Moreover, it is compared primarily with models trained from scratch, leaving open how a frozen off-the-shelf natural-image-based representation fares against fMRI foundation models.

In this work, we therefore ask:
\emph{Can competitive fMRI-based prediction be achieved without any fMRI-specific pre-training?}
To answer this, we present \textbf{{\ours}}, which builds fMRI representations from a DCAE pre-trained exclusively on natural images and kept frozen throughout, and adapts only the readout to the structure of the downstream task. 
Our study spans both trait and state prediction, which differ in
whether the target is constant or varies within a subject, respectively.
We first consider resting-state fMRI-based trait prediction.
Because the target is constant over time, we summarize each fMRI sequence using the temporal mean and log-standard deviation of its frame-level DCAE representations and fit a linear predictor.
Despite its simplicity, this approach generally outperforms existing fMRI foundation models under linear probing.
Fusing the predictions obtained under two complementary normalization schemes goes further, remaining competitive on most tasks even when those are fully fine-tuned.
These results show that strong trait prediction can be achieved without fMRI-specific pre-training or end-to-end encoder adaptation. 

We next consider task-fMRI-based state prediction, where the target changes within a subject.
Here, the same order-invariant aggregation performs
substantially worse, indicating that temporal information is important for state decoding.
We therefore represent each frame as a single token and model their temporal dependencies with a shallow Transformer while keeping the DCAE frozen.
On cognitive task-state decoding~\citep{HCPTask} and object-category decoding~\citep{NSD}, this simple temporal readout perform comparably to the strongest fMRI foundation models. It also outperforms TABLeT while using a 27-fold shorter attention sequence (one token per frame) and 40\% fewer trainable parameters.


We further examine why such a frozen off-the-shelf representation supports downstream tasks. Our analyses show that a substantial fraction of fMRI variation is captured by the principal directions of the DCAE latent space, and that localized perturbations injected into resting-state fMRI are recovered more accurately from the frozen DCAE features than from the evaluated foundation-model representations.
Together, these results show that the performance of the fMRI foundation models on current popular benchmarks can be matched by a frozen natural-image autoencoder, paired with a task-appropriate readout. We therefore argue that frozen natural-image representations should serve as a baseline for assessing the added value of fMRI-specific pre-training.

\section{Preliminaries}
\label{sec:preliminaries}

\subsection{fMRI Data Types \& Normalization Schemes}
\label{subsec:datatypes}

Existing fMRI foundation models consider three fMRI data types:
\textit{volume}, \textit{cortical flat map}, and \textit{parcellation},
which we denote by the superscript $r\in\{V,F,P\}$.
A volumetric fMRI sequence is denoted by
$\mathbf{X}^{V}\in\mathbb{R}^{T\times1\times D\times H\times W}$, where $T$ is the number of temporal frames and $D$, $H$, and $W$ are the spatial dimensions of the volume, and a cortical flat-map sequence by
$\mathbf{X}^{F}\in\mathbb{R}^{T\times1\times H_F\times W_F}$, where
$H_F$ and $W_F$ are the spatial dimensions of the flat map.
Both retain spatial structure without aggregation over predefined
regions, whereas a parcellation
$\mathbf{X}^{P}\in\mathbb{R}^{T\times N_P}$ averages voxel signals
within $N_P$ predefined regions of interest.
Our {\ours} operates on $\mathbf{X}^{V}$ and $\mathbf{X}^{F}$, since both
retain the spatial structure required by a 2D image encoder; we consider
$\mathbf{X}^{P}$ only through the baselines.

For $\mathbf{X}^{V}$ and $\mathbf{X}^{F}$, we introduce three
normalization schemes.
\textit{Global normalization} computes a shared set of statistics over
the entire fMRI sequence and applies the same transformation to all
temporal frames and spatial coordinates, as commonly adopted by existing
fMRI foundation models \citep{swift, neurostorm, omnifmri}.
\textit{Coordinate-wise normalization} independently standardizes the
temporal signal at each spatial coordinate, whereas
\textit{frame-wise normalization} independently standardizes the spatial
values within each temporal frame.
In our experiments, we consider two normalization schemes:
global normalization and \textit{coordinate--frame normalization}, where
the latter sequentially applies coordinate-wise normalization followed by
the frame-wise normalization.
Throughout this paper, we use normalization subscripts only for
$\mathbf{X}^{V}$ and $\mathbf{X}^{F}$.
We index this choice by $n\in\{g,cf\}$ and write $\mathbf{X}^{r}_{n}$
for $r\in\{V,F\}$, where $\mathbf{X}^{r}_{g}$ is globally normalized and
$\mathbf{X}^{r}_{cf}$ coordinate--frame-normalized; for
$\mathbf{X}^{P}$ we follow each baseline model's own convention.

\subsection{fMRI Foundation Models \& Downstream Tasks}
\label{subsec:fm_tasks}

\paragraph{fMRI foundation models.}
Following the data types in
\Cref{subsec:datatypes}, existing fMRI foundation models can be grouped
into volume-, cortical flat-map-, and parcellation-based approaches.
Volume-based models such as SwiFT \citep{swift}, NeuroSTORM \citep{neurostorm}, and Omni-fMRI \citep{omnifmri} directly
process volumetric fMRI.
CortexMAE \citep{cortexmae} instead operates on cortical flat maps.
Parcellation-based models, including BrainMASS \citep{brainmass}, Brain-JEPA \citep{brainjepa}, BrainHarmonix-F \citep{brainharmony}, and Brain-DiT \citep{braindit}, operate on ROI-level representations.
These models are generally pre-trained on large-scale neuroimaging
datasets with self-supervised objectives.

\paragraph{Downstream tasks.}
We broadly categorize downstream tasks into \textit{trait prediction}
and \textit{state prediction}, following \citet{cortexmae}, depending on whether the target varies
within a subject.
Trait prediction targets subject-level attributes that remain fixed
within a subject, such as age or sex.
Since the target does not vary over time, such tasks may be addressed
without explicitly modeling temporal dependencies, and representations
that capture stable differences between subjects are expected to be
informative.
State prediction, on the contrary, targets cognitive or
stimulus-driven states, such as labels for cognitive tasks \citep{HCPTask} or viewed stimuli \citep{NSD}.
Here the label changes within a subject, so modeling temporal variations is likely to be important.

\section{\ours}
\label{sec:method}

\begin{figure}[t]
    \centering
    \includegraphics[width=0.85\linewidth, height=6.5cm]{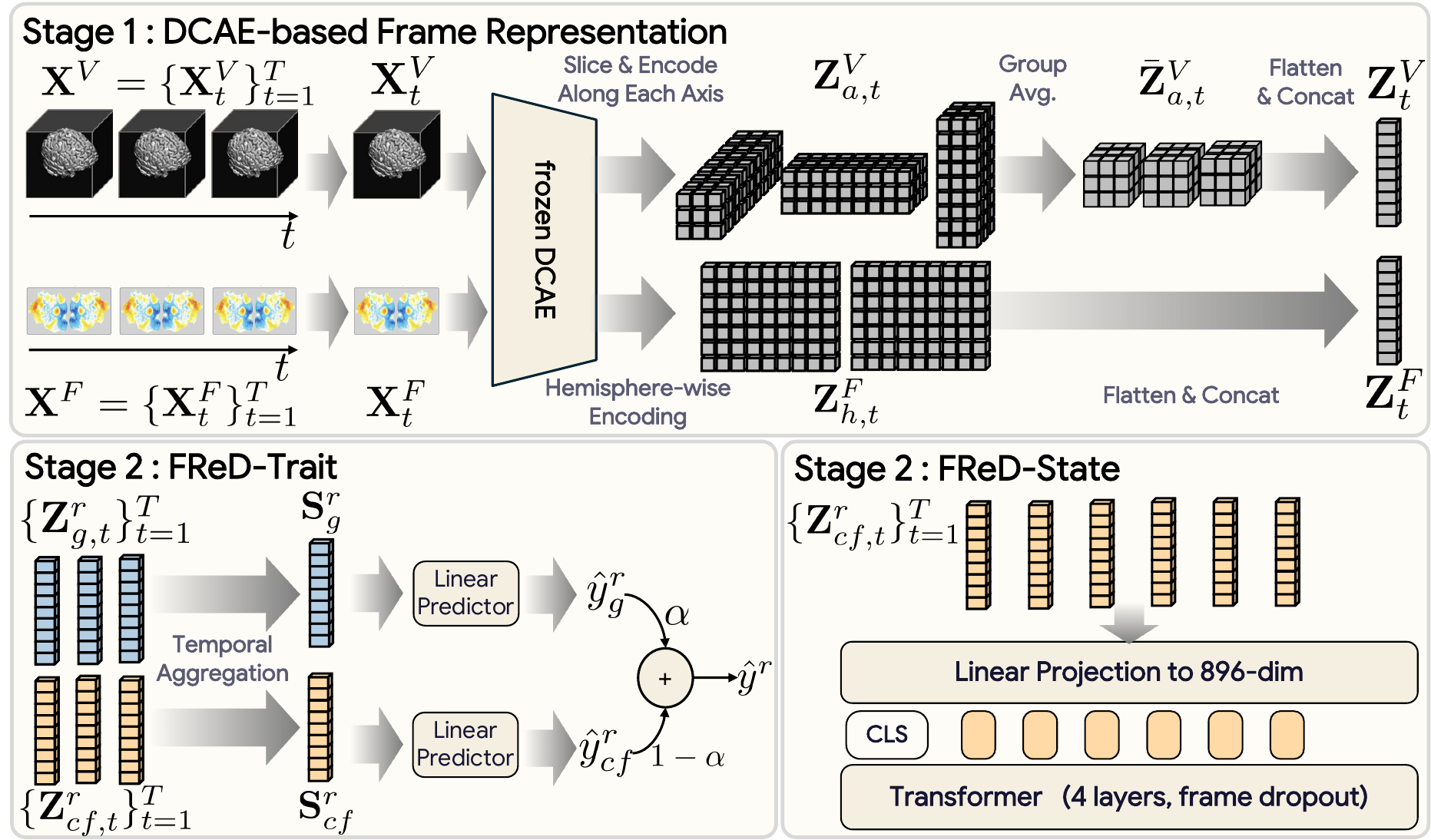}
    \vspace{-0.1in}
\caption{
    Overview of {\ours}.
    A frozen DCAE pre-trained on natural images constructs a frame
    representation for each fMRI frame, for both $\mathbf{X}^{V}$ and
    $\mathbf{X}^{F}$.
    For trait prediction, the frame representations are
    summarized by their temporal mean and log-standard deviation and
    read out by a linear predictor, and the predictions obtained under
    the two normalization schemes $n\in\{g,cf\}$ are combined by late
    fusion.
    For state prediction, the frame representations obtained
    under coordinate--frame normalization are projected into a common
    embedding space and processed by a Transformer.
}
    \label{fig:main_figure}
\vspace{-0.2in}
\end{figure}

We present \textbf{\ours}, a framework for constructing and
modeling \textbf{f}MRI \textbf{Re}presentations based on a natural-image-pre-trained \textbf{D}CAE.
{\ours} operates on two fMRI data types, $\mathbf{X}^V$ and
$\mathbf{X}^F$, and consists of two stages, as illustrated in
\Cref{fig:main_figure}: DCAE-based frame representation and task-dependent readout.

Let
$
    \mathbf{X}^{r}_{n}
    =
    \{\mathbf{X}^{r}_{n,t}\}_{t=1}^{T},
    r\in\{V,F\},
    n\in\{g,cf\},
$
denote an fMRI sequence of representation type $r$ normalized by scheme
$n$. A frozen DCAE
encoder first constructs a representation for each temporal frame
$\mathbf{X}^{r}_{n,t}$. {\ours} then processes the resulting frame
sequence according to the downstream task. For trait prediction, it
summarizes the temporal distribution of the frame representations and
fits a linear predictor. 
For state prediction, it collapses each frame to a single token before the Transformer, so that spatial mixing happens in the projection and the Transformer models only temporal structure. This contrasts with TABLeT~\citep{tablet}, which retains 27 tokens per frame and relies on self-attention to model both spatial and temporal relationships.

\subsection{DCAE-based Frame Representation}
\label{subsec:frame_repr}

The DCAE~\citep{dcae} is a convolutional autoencoder designed to provide highly compressed latent spaces for high-resolution image generation. We use the publicly released \texttt{dc-ae-f32c32-in-1.0} checkpoint without any modification, which downsamples an input image by a factor of 32 along each spatial dimension into a 32-channel latent map.
The checkpoint is trained only on ImageNet \citep{imagenet}, and we keep it frozen throughout.

\paragraph{Frame-wise encoding.}
We independently encode each fMRI frame with the DCAE.
The same encoder is shared across data types, normalization schemes, and temporal frames.
Since the encoding procedure is identical for both normalization
schemes, we omit the subscript $n$ throughout this subsection and write
$\mathbf{X}^{r}_{t}$ for a single frame.

\paragraph{$\mathbf{X}^{V}$ case.}
We follow TABLeT in slicing the volume along the three anatomical axes and encoding each slice independently with the frozen DCAE, and depart from it in how the resulting latents are assembled into a frame representation.
A volumetric fMRI frame is represented as
$
    \mathbf{X}^{V}_{t}
    \in
    \mathbb{R}^{1\times D\times H\times W},
$
where $D=H=W=96$ in our experiments.
We slice the volume along the three anatomical axes
$a\in\mathcal{A}=\{\mathrm{sag},\mathrm{cor},\mathrm{axi}\}$,
which yields $N_{\mathrm{sag}}=D$, $N_{\mathrm{cor}}=H$, and
$N_{\mathrm{axi}}=W$ two-dimensional slices, respectively.
Each axis thus produces 96 slices of size $96\times96$ in our
implementation.
Each slice is replicated across three channels and independently encoded
by the frozen DCAE, yielding
$    
    \mathbf{Z}^{V}_{a,t}
    \in
    \mathbb{R}^{N_a\times C_z\times H_z\times W_z},
$
where $C_z\times H_z\times W_z$ denotes the DCAE latent size per slice,
which is $32\times3\times3$.

We then partition the $N_a$ slices along each axis into $G$ contiguous,
equally sized groups and average the DCAE latents within each group,
which gives
$
    \bar{\mathbf{Z}}^{V}_{a,t}
    \in
    \mathbb{R}^{G\times C_z\times H_z\times W_z}.
$
Without this step a frame representation would have $3N_aC_zH_zW_z=82{,}944$ dimensions, which is impractical both for a linear probe and for a linear projection into the Transformer; grouping reduces this to $3GC_zH_zW_z$ while retaining spatial granularity along each axis.
Finally, we flatten and concatenate the grouped latents from the three
axes to construct the frame representation:
\begin{equation}
    \mathbf{Z}^{V}_{t}
    =
    \operatorname{Concat}_{a\in\mathcal{A}}
    \left(
        \operatorname{Flatten}
        \left(
            \bar{\mathbf{Z}}^{V}_{a,t}
        \right)
    \right)
    \in
    \mathbb{R}^{3G C_z H_z W_z}.
\label{eq:volume_frame_latent}
\end{equation}
TABLeT instead keeps the slice and latent-grid positions as 27 separate tokens per frame; grouping and flattening here reduce this to one, which is the change we ablate in \Cref{tab:tablet_v1_vs_v2}.
We use $G=3$ for trait prediction and $G=24$ for state prediction, supported by the sensitivity analysis in \Cref{fig:ablations}.

\paragraph{$\mathbf{X}^{F}$ case.}
A cortical flat-map frame is represented as
$
    \mathbf{X}^{F}_{t}
    \in
    \mathbb{R}^{1\times H_F\times W_F},
$
where $(H_F, W_F) = (224, 576)$ in our experiments.
We split the frame along its width into the left- and right-hemisphere
maps $h\in\mathcal{H}=\{\mathrm{lh},\mathrm{rh}\}$, each of size
$H_F\times(W_F/2)$.
Each hemisphere map is replicated across three channels and
independently encoded by the frozen DCAE,
yielding
$
    \mathbf{Z}^{F}_{h,t}
    \in
    \mathbb{R}^{C_z\times H_z\times W_z},
$
where each hemisphere map has size $H_F\times(W_F/2)=224\times288$, and the DCAE downsamples each spatial dimension by 32, giving $C_z\times H_z\times W_z=32\times7\times9$.
We then flatten and concatenate the latents of the two hemispheres to
construct the frame representation:
\begin{equation}
    \mathbf{Z}^{F}_{t}
    =
    \operatorname{Concat}_{h\in\mathcal{H}}
    \left(
        \operatorname{Flatten}
        \left(
            \mathbf{Z}^{F}_{h,t}
        \right)
    \right)
    \in
    \mathbb{R}^{2C_zH_zW_z},
\label{eq:flat_map_frame_latent}
\end{equation}
which has dimension $4{,}032$ in our implementation.
Unlike the volumetric case, there is no slice axis to aggregate, and its dimensionality is moderate compared with that of $X^V$. The same flat-map representation is therefore used for both trait and state prediction.

\subsection{Task-dependent Readout}
\label{subsec:task_modeling}
We hereafter write
$\mathbf{Z}^{r}_{n,t}$ for the frame representation obtained from
$\mathbf{X}^{r}_{n,t}$. Starting from the frame representations $\mathbf{Z}^{r}_{n,t}$, {\ours}
uses statistical aggregation with linear probing for trait prediction
({\oursl}) and Transformer-based temporal modeling for state prediction
({\ourst}).
The two differ in how they model temporal information and which normalization schemes they use.

\paragraph{\oursl.}
Since trait labels do not vary over time, we compress the
temporal dynamics into the element-wise mean and log-standard deviation:
\begin{equation}
    \mathbf{S}^{r}_{n}
    =
    \operatorname{Concat}
    \left(
        \frac{1}{T}\sum_{t=1}^{T}\mathbf{Z}^{r}_{n,t},
        \;
        \log\left(
            \operatorname{Std}_{t}(\mathbf{Z}^{r}_{n,t})
            +\epsilon
        \right)
    \right),
\end{equation}
where $\operatorname{Std}_t$ denotes the element-wise standard deviation
over the $T$ frames and $\epsilon$ is a small constant for numerical
stability.
A linear predictor is then fitted on $\mathbf{S}^{r}_{n}$.
As the two normalization schemes are expected to retain complementary
information, we apply this pipeline independently to
$\mathbf{X}^{r}_{g}$ and $\mathbf{X}^{r}_{cf}$, and combine the
resulting predictions $\hat{y}^{r}_{g}$ and $\hat{y}^{r}_{cf}$ by
prediction-level late fusion:
\begin{equation}
    \hat{y}^{r}
    =
    \alpha\,\hat{y}^{r}_{g}
    +
    (1-\alpha)\,\hat{y}^{r}_{cf},
    \qquad
    \alpha\in[0,1],
\end{equation}
where $\alpha$ is selected based on the validation set, detailed in \Cref{subsec:train_eval_details}.
\paragraph{\ourst.} 
For state prediction, we explicitly model the temporal dependencies among frame representations. Each $\mathbf{Z}^{r}_{cf,t}$ is projected into an 896-dimensional embedding by a learnable linear layer, and we prepend a learnable \texttt{[CLS]} token to the resulting sequence. The sequence is processed by a four-layer self-attention Transformer with 1D-RoPE~\citep{rope}. The backbone follows modern practices, including GQA~\citep{gqa} and Q/K normalization~\citep{qknorm} with soft-capped attention logits~\citep{soft_cap}. We give the full specification in \Cref{tab:tablet_v1_vs_v2_arch}. The output at the \texttt{[CLS]} token is used for prediction. 
We additionally apply frame dropout during training as regularization. 
In contrast to {\oursl}, we use only coordinate--frame-normalized inputs; we examine this choice empirically in \Cref{fig:hbn_results}.
\section{Experimental Results}
\label{sec:experimental results}

\subsection{Experimental Settings}
\label{subsec:experimental_settings}
\paragraph{Datasets and tasks.}
We evaluate trait prediction on four resting-state fMRI datasets:
ADHD classification using ADHD-200~\citep{ADHD}, autism spectrum
disorder (ASD) classification using ABIDE-II~\citep{ABIDE2}, mild
cognitive impairment (MCI) classification using ADNI~\citep{ADNI}, and
prediction of sex, age, and intelligence using HCP-A~\citep{HCPA}.
For state prediction, we consider three task-fMRI datasets: viewed-movie binary classification on HBN~\citep{HBN}, 21-class
cognitive task-state decoding on HCP-Task~\citep{HCPTask}, and decoding
of 24 MS-COCO object categories from NSD~\citep{NSD}.
All splits are constructed at the subject level, and we use the fixed
splits provided by BrainMarks~\citep{cortexmae} for HCP-Task and NSD. More details are provided in \Cref{subsec:datasets_detail}.

\paragraph{Evaluation protocol.}
The linear-probing experiments in \Cref{fig:lp_results}, \Cref{tab:hbn_movie_gender_lp}, and \Cref{fig:lp_rp768_results} use 500 independently sampled 85/15 train--test splits.
All other experiments use 70/15/15 train/validation/test splits, 12 splits for the clinical diagnosis tasks, and four splits for HCP-A and HBN; HCP-Task and NSD instead use three random seeds on their fixed splits.
We report accuracy (Acc.), AUROC (AUC), and F1 for classification, and mean absolute error (MAE) and Pearson's correlation ($\rho$) for regression, with MAE expressed in units of the standardized target.

\paragraph{Baselines.}
We compare {\ours} against eight fMRI foundation models and
TABLeT~\citep{tablet}: BrainMASS~\citep{brainmass},
Brain-JEPA~\citep{brainjepa}, BrainHarmonix-F (BH-F)~\citep{brainharmony}, and
Brain-DiT~\citep{braindit} operate on parcellations;
SwiFT~\citep{swift}, NeuroSTORM~\citep{neurostorm}, and
Omni-fMRI~\citep{omnifmri} on volumes; and CortexMAE~\citep{cortexmae}
on cortical flat maps.
Since BrainMASS operates on functional connectivity rather than on the fMRI time series, we report it as a reference line in \Cref{fig:lp_results,fig:hbn_results} indicating the performance attainable from a basic connectivity-based representation.
TABLeT is the most closely related baseline to ours, as it also encodes 2D fMRI slices using a DCAE pre-trained on natural images, but it uses a single data type and modeling scheme across tasks.

\subsection{Trait Prediction}
\label{subsec:trait_prediction}

\begin{figure}[t]
    \centering
    \includegraphics[width=\linewidth]{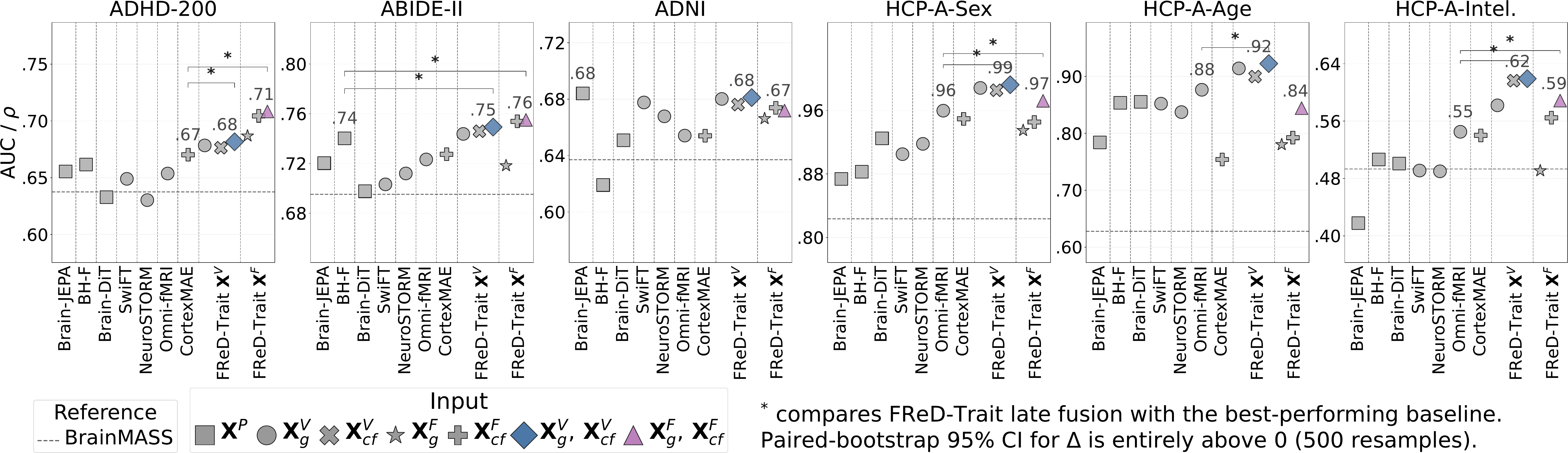}
    \vspace{-0.25in}
\caption{
Linear-probing results on trait prediction tasks over 500 random splits, 
reported as AUC/$\rho$.
}
    \label{fig:lp_results}
\vspace{-0.2in}
    
\end{figure}

\definecolor{lprow}{HTML}{FFF6CC}
\newcommand{\rowstrut}{\rule[-1.0ex]{0pt}{3.2ex}}

\begin{table*}[!ht]
\caption{
Comparison of trait prediction performance. All foundation models are fully fine-tuned, TABLeT is trained from scratch, and {\oursl} uses linear probing. Results are reported as the mean and standard deviation across data splits. \textbf{Bold} and \underline{underlined} values indicate the best and second-best performance for each metric, respectively.
}
\vspace{-0.1in}
\centering
\renewcommand{\arraystretch}{1.25}
\resizebox{\linewidth}{!}{%
\begin{tabular}{l|c|cc|cc|cc|cc|cc|cc}
\hline
\multirow{3}{*}{Method} & \multirow{3}{*}{Input}
& \multicolumn{2}{c|}{ADHD-200}
& \multicolumn{2}{c|}{ABIDE-II}
& \multicolumn{2}{c|}{ADNI}
& \multicolumn{6}{c}{HCP-A} \\ \cline{3-14}
 & & \multicolumn{2}{c|}{Diagnosis}
 & \multicolumn{2}{c|}{Diagnosis}
 & \multicolumn{2}{c|}{Diagnosis}
 & \multicolumn{2}{c|}{Sex}
 & \multicolumn{2}{c|}{Age}
 & \multicolumn{2}{c}{Intelligence} \\
 & & AUC $\uparrow$ & \multicolumn{1}{c|}{F1 $\uparrow$}
 & AUC $\uparrow$ & \multicolumn{1}{c|}{F1 $\uparrow$}
 & AUC $\uparrow$ & \multicolumn{1}{c|}{F1 $\uparrow$}
 & AUC $\uparrow$ & \multicolumn{1}{c|}{F1 $\uparrow$}
 & MAE $\downarrow$ & \multicolumn{1}{c|}{$\rho$ $\uparrow$}
 & MAE $\downarrow$ & $\rho$ $\uparrow$ \\ \hline

Brain-JEPA & $\mathbf{X}^P$ \rowstrut
& \meanstd{.660}{.00} & \meanstd{.606}{.01}
& \meanstd{.720}{.05} & \meanstd{.622}{.04}
& \meanstd{\underline{.693}}{.07} & \meanstd{.597}{.05}
& \meanstd{.917}{.00} & \meanstd{.829}{.00}
& \meanstd{.418}{.00} & \meanstd{.849}{.00}
& \meanstd{.681}{.01} & \meanstd{.487}{.01} \\

BH-F & $\mathbf{X}^P$ \rowstrut
& \meanstd{.631}{.05} & \meanstd{.573}{.06}
& \meanstd{.697}{.03} & \meanstd{.592}{.05}
& \meanstd{.601}{.06} & \meanstd{.452}{.12}
& \meanstd{.867}{.01} & \meanstd{.729}{.10}
& \meanstd{.435}{.02} & \meanstd{.838}{.02}
& \meanstd{.679}{.04} & \meanstd{.498}{.06} \\

Brain-DiT & $\mathbf{X}^P$ \rowstrut
& \meanstd{.648}{.05} & \meanstd{.568}{.04}
& \meanstd{.712}{.04} & \meanstd{.535}{.08}
& \meanstd{.659}{.05} & \meanstd{.484}{.14}
& \meanstd{.923}{.02} & \meanstd{.833}{.03}
& \meanstd{.386}{.01} & \meanstd{.877}{.00}
& \meanstd{.674}{.01} & \meanstd{.546}{.01} \\ \hline

SwiFT & $\mathbf{X}_g^V$ \rowstrut
& \meanstd{.664}{.05} & \meanstd{\textbf{.633}}{.05}
& \meanstd{.711}{.03} & \meanstd{.606}{.04}
& \meanstd{.688}{.07} & \meanstd{.582}{.06}
& \meanstd{.946}{.01} & \meanstd{.853}{.02}
& \meanstd{.350}{.03} & \meanstd{.897}{.02}
& \meanstd{.678}{.03} & \meanstd{.507}{.02} \\

NeuroSTORM & $\mathbf{X}_g^V$ \rowstrut
& \meanstd{.658}{.04} & \meanstd{.572}{.07}
& \meanstd{.716}{.03} & \meanstd{.606}{.02}
& \meanstd{.673}{.04} & \meanstd{\textbf{.607}}{.02}
& \meanstd{.974}{.01} & \meanstd{.888}{.02}
& \meanstd{\underline{.343}}{.02} & \meanstd{\underline{.906}}{.01}
& \meanstd{.670}{.04} & \meanstd{.525}{.05} \\

Omni-fMRI & $\mathbf{X}_g^V$ \rowstrut
& \meanstd{.676}{.05} & \meanstd{.598}{.05}
& \meanstd{.712}{.03} & \meanstd{.586}{.03}
& \meanstd{.681}{.06} & \meanstd{.584}{.06}
& \meanstd{\underline{.975}}{.01} & \meanstd{\underline{.910}}{.01}
& \meanstd{.351}{.01} & \meanstd{.894}{.01}
& \meanstd{.657}{.04} & \meanstd{.546}{.05} \\ \hline

CortexMAE & $\mathbf{X}_{cf}^F$ \rowstrut
& \meanstd{\underline{.690}}{.05} & \meanstd{.605}{.06}
& \meanstd{.735}{.03} & \meanstd{\textbf{.639}}{.04}
& \meanstd{.638}{.05} & \meanstd{.540}{.06}
& \meanstd{.944}{.02} & \meanstd{.859}{.02}
& \meanstd{.508}{.02} & \meanstd{.774}{.02}
& \meanstd{.661}{.05} & \meanstd{.539}{.06} \\ \hline

TABLeT & $\mathbf{X}_g^V$ \rowstrut
& \meanstd{.639}{.04} & \meanstd{.583}{.06}
& \meanstd{.713}{.03} & \meanstd{.624}{.04}
& \meanstd{\textbf{.702}}{.05} & \meanstd{\underline{.602}}{.05}
& \meanstd{.963}{.02} & \meanstd{.872}{.05}
& \meanstd{.362}{.02} & \meanstd{.891}{.01}
& \meanstd{.660}{.04} & \meanstd{.549}{.03} \\

TABLeT & $\mathbf{X}_{cf}^V$ \rowstrut
& \meanstd{.624}{.05} & \meanstd{.592}{.06}
& \meanstd{.705}{.03} & \meanstd{.588}{.04}
& \meanstd{.620}{.07} & \meanstd{.564}{.06}
& \meanstd{.946}{.02} & \meanstd{.860}{.04}
& \meanstd{.398}{.01} & \meanstd{.871}{.01}
& \meanstd{.667}{.05} & \meanstd{.513}{.04} \\ \hline

\rowcolor{lprow}
\oursl & $\mathbf{X}_{g}^V, \mathbf{X}_{cf}^V$ \rowstrut
& \meanstd{.686}{.05} & \meanstd{.604}{.05}
& \meanstd{\underline{.743}}{.03} & \meanstd{\underline{.629}}{.04}
& \meanstd{.690}{.04} & \meanstd{.591}{.05}
& \meanstd{\textbf{.995}}{.00} & \meanstd{\textbf{.956}}{.03}
& \meanstd{\textbf{.320}}{.01} & \meanstd{\textbf{.917}}{.01}
& \meanstd{\textbf{.633}}{.06} & \meanstd{\textbf{.607}}{.05} \\

\rowcolor{lprow}
\oursl & $\mathbf{X}_{g}^F, \mathbf{X}_{cf}^F$ \rowstrut
& \meanstd{\textbf{.717}}{.03} & \meanstd{\underline{.607}}{.05}
& \meanstd{\textbf{.747}}{.03} & \meanstd{.615}{.05}
& \meanstd{.679}{.05} & \meanstd{.541}{.04}
& \meanstd{.969}{.00} & \meanstd{.884}{.00}
& \meanstd{.476}{.02} & \meanstd{.828}{.01}
& \meanstd{\underline{.655}}{.03} & \meanstd{\underline{.576}}{.04} \\
\hline
\end{tabular}}
\label{tab:resting_ft_results}
\vspace{-0.25in}
\end{table*}

\paragraph{Linear probing.}
We first compare the downstream utility of frozen representations.
We standardize the input features and perform five-fold cross-validation, using \texttt{LogisticRegressionCV} from scikit-learn \citep{scikit} for classification and \texttt{RidgeCV} for regression.

As shown in \Cref{fig:lp_results}, FReD-Trait outperforms the evaluated foundation-model baselines on five of the six tasks under linear evaluation,
with ADNI being the only case where the methods perform comparably.
These results show that frozen DCAE features, summarized by simple temporal statistics under a single normalization scheme, can support strong trait prediction without fMRI-specific pre-training.
Late fusion of the predictions obtained under the two normalization schemes (blue diamond and pink triangle) further improves performance generally, indicating that
$\mathbf{X}^{r}_{g}$ and $\mathbf{X}^{r}_{cf}$ retain complementary information relevant to trait prediction.
The selected fusion weights vary across tasks, suggesting that the
preferred normalization scheme is itself task-dependent; we report
their distribution in \Cref{subsec:fusion_weights}.
One may argue that this comparison is confounded by the dimensionality of the representations, which is larger for {\oursl}
than for the foundation models.
To address this, we adopt dimension reduction and find its performance to remain robust, as detailed in \Cref{subsec:dim_matched_lp}.

\paragraph{Can fine-tuning yield better performance than \oursl?}
We next investigate whether fine-tuning existing fMRI foundation models
on individual downstream tasks improves their performance.
For each foundation model, we follow its original fine-tuning protocol
and search at least eight learning rates per dataset, with the full
search ranges reported in \Cref{sec:experimental_details}.
In addition, we evaluate partial fine-tuning, in
which only the final few layers of the pre-trained encoder are updated.

The full fine-tuning results are reported in Tab.~\ref{tab:resting_ft_results},
and the partial fine-tuning results in Sec.~\ref{subsec:partial_ft}.
Despite adapting the entire pre-trained encoder to each downstream
task, the fully fine-tuned foundation models remain broadly
competitive with FReD-Trait rather than consistently surpassing it.
Moreover, we additionally observe that fine-tuning itself does not yield consistent improvements over linear probing on the matched splits, as detailed in \Cref{subsec:fixed_split}.
These results show that
a frozen off-the-shelf representation paired with simple temporal aggregation can provide a competitive alternative to task-specific adaptation of fMRI foundation models.
\subsection{State Prediction}
\label{subsec:state_prediction}

\definecolor{lprow}{HTML}{FFF6CC}

\begin{table*}[!ht]
\centering
\begin{minipage}[t]{0.545\textwidth}
  \vspace{0pt}
  \centering
  \includegraphics[width=\linewidth]{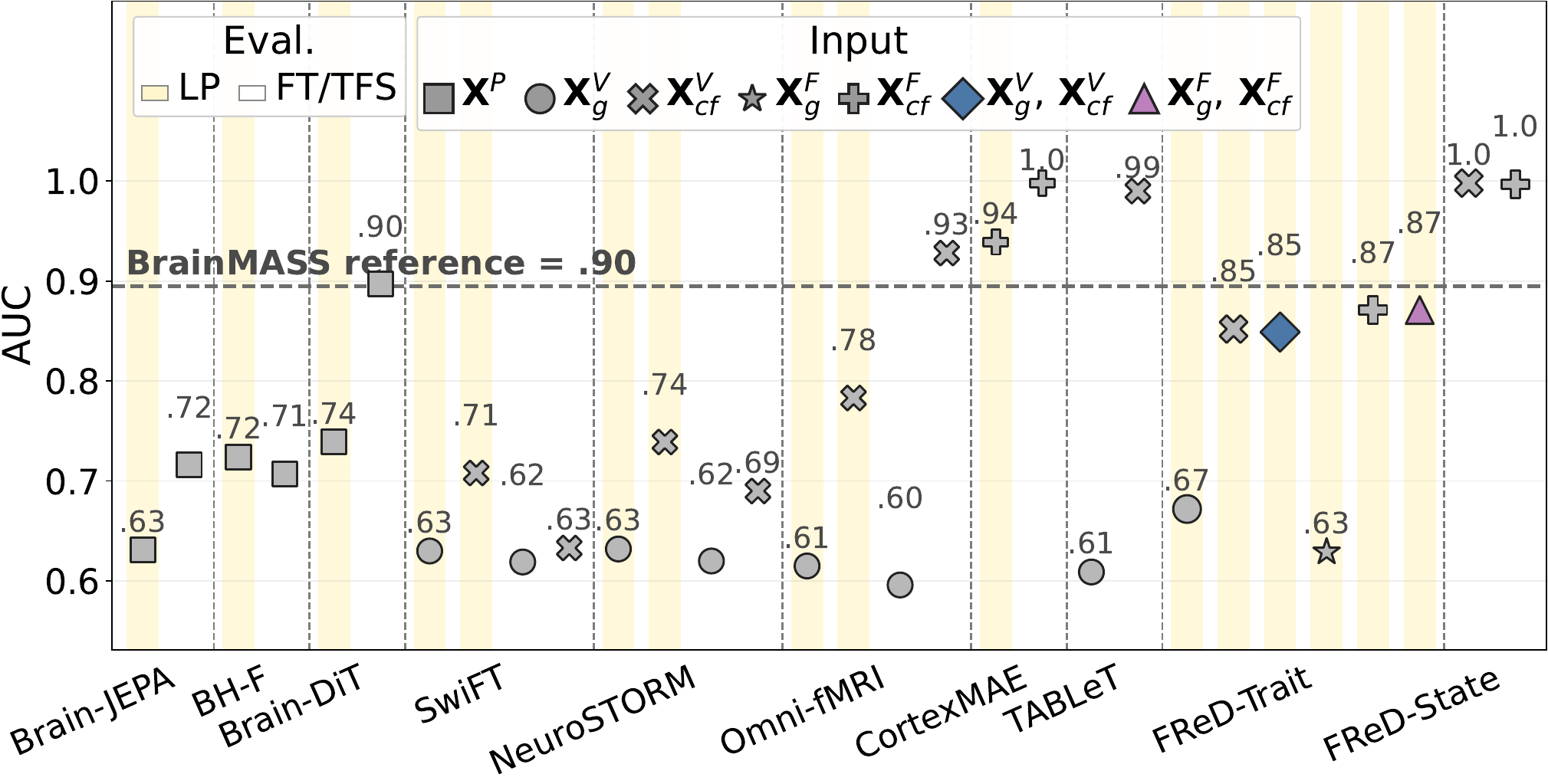}
  \vspace{-0.3in}
  \captionof{figure}{HBN-Movie task results with four splits.}
  \label{fig:hbn_results}
\end{minipage}\hfill
\begin{minipage}[t]{0.445\textwidth}
  \centering
  \captionof{table}{Results on state prediction tasks. \textbf{Bold} and \underline{underlined} values indicate the best and second-best performance, respectively. FT: Full fine-tuning, TFS: Train from scratch.}
  \vspace{-0.1in}
  \label{tab:task_fmri_results}
  \renewcommand{\arraystretch}{1.25}
  \resizebox{\linewidth}{!}{%
  \begin{tabular}{l|cc|cc|c|c}
  \hline
  \multirow{3}{*}{Method} & \multirow{3}{*}{Input} & \multirow{3}{*}{Eval}
   & \multicolumn{2}{c|}{\multirow{2}{*}{HBN-Movie}}
   & \multicolumn{1}{c|}{\multirow{2}{*}{HCP-Task}}
   & \multicolumn{1}{c}{\multirow{2}{*}{NSD}} \\
   & & & \multicolumn{2}{c|}{} & & \\ \cline{4-7}
   & & & AUC $\uparrow$ & F1 $\uparrow$ & Acc. $\uparrow$ & Acc. $\uparrow$ \\ \hline
  \rowcolor{white}
  Omni-fMRI & $\mathbf{X}_{cf}^{V}$ & \rowstrut FT
    & \meanstd{.928}{.03} & \meanstd{.838}{.02} & \meanstd{94.5}{0.2} & \meanstd{21.5}{0.9}\\
  \rowcolor{white}
  CortexMAE & $\mathbf{X}_{cf}^{F}$ & \rowstrut FT
    & \meanstd{\underline{.997}}{.00} & \meanstd{.959}{.01}
    & \meanstd{\textbf{99.2}}{0.1} & \meanstd{\underline{29.8}}{0.5} \\
  \rowcolor{white}
  TABLeT & $\mathbf{X}_{cf}^{V}$ & \rowstrut TFS
    & \meanstd{.990}{.01} & \meanstd{.948}{.03}
    & \meanstd{94.2}{0.2} & \meanstd{23.9}{1.0} \\ \hline
  \rowcolor{white}
  \ourst & $\mathbf{X}_{cf}^{V}$ & \rowstrut TFS
    & \meanstd{\textbf{.998}}{.00} & \meanstd{\textbf{.973}}{.01}
    & \meanstd{97.4}{0.1} & \meanstd{29.5}{0.2} \\
  \rowcolor{white}
  \ourst & $\mathbf{X}_{cf}^{F}$ & \rowstrut TFS
    & \meanstd{\underline{.997}}{.00} & \meanstd{\underline{.971}}{.01}
    & \meanstd{\underline{98.8}}{0.2} & \meanstd{\textbf{30.0}}{0.6} \\
  \hline\hline
  \end{tabular}}
\end{minipage}
\vspace{-0.2in}
\end{table*}
\paragraph{Insights from the HBN-Movie task.}

We first consider a relatively simple state prediction task, where the model predicts which of two movies a participant was watching from the corresponding fMRI sequence.
As shown in \Cref{fig:hbn_results}, the volume-based foundation
models, SwiFT, NeuroSTORM, and Omni-fMRI, perform only slightly above
chance on $\mathbf{X}^{V}_{g}$ under both linear probing and full
fine-tuning, whereas linear probing on $\mathbf{X}^{V}_{cf}$
consistently yields better performance.
We further find that SwiFT and NeuroSTORM do not substantially improve
even when fully fine-tuned on $\mathbf{X}^{V}_{cf}$, and that among the
volume-based models only Omni-fMRI benefits from doing so.
The effect is even more pronounced for the DCAE-based models, TABLeT and {\oursl}, whose performance improves substantially on coordinate-frame-normalized inputs.
Coordinate-frame normalization is thus critical for state prediction, which motivates its exclusive use in {\ourst}.

Although {\oursl} performs strongly on trait prediction, it
underperforms TABLeT on HBN-Movie.
We attribute this to the fact that {\oursl} compresses an entire fMRI sequence into its temporal mean and log-standard deviation and thus discards the temporal ordering of the frame representations, which may help distinguishing stimulus-induced states in this task.
Statistical temporal aggregation with linear probing is therefore insufficient to capture within-subject variability, which motivates the temporal modeling in {\ourst}.
By explicitly modeling the temporal interactions among frame
representations, {\ourst} outperforms {\oursl} by a large margin and
becomes competitive with the strongest foundation models, indicating
that DCAE latents retain sufficient information for state prediction
once paired with an appropriate temporal architecture.

\paragraph{More challenging tasks.}
We next consider two harder state prediction tasks, 21-class cognitive task-state decoding on HCP-Task and 24-class object-category decoding on NSD.
For these evaluations, we include Omni-fMRI and CortexMAE that clearly outperformed BrainMASS \citep{brainmass} on HBN-Movie, along with TABLeT, the most closely related DCAE-based baseline.
As shown in \Cref{tab:task_fmri_results}, {\ourst} outperforms Omni-fMRI and TABLeT by clear margins on both tasks, while performing comparably to CortexMAE.
Between the two data types, $\mathbf{X}^{F}$ performs better on HCP-Task and comparably on NSD. With $\mathbf{X}^{F}$, {\ourst} matches the strongest flat-map baseline.

\begin{figure}[t]
    \centering
    \begin{minipage}[b]{0.495\linewidth}
        \centering
        \includegraphics[width=\linewidth]{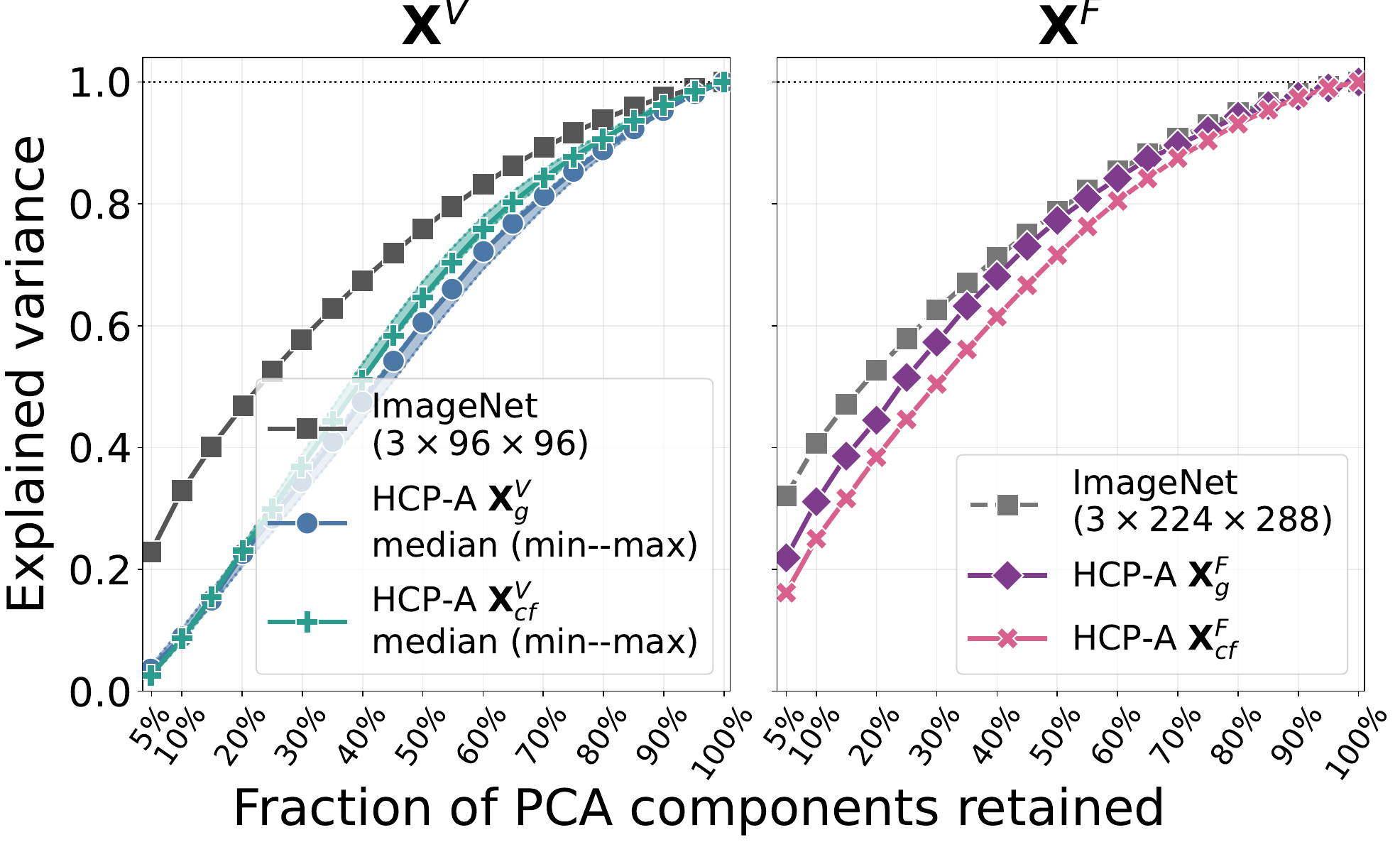}
        \\[1pt]{\small (a)} \vspace{-0.15in}
    \end{minipage}
    \hfill
    \begin{minipage}[b]{0.495\linewidth}
        \centering
        \includegraphics[width=\linewidth]{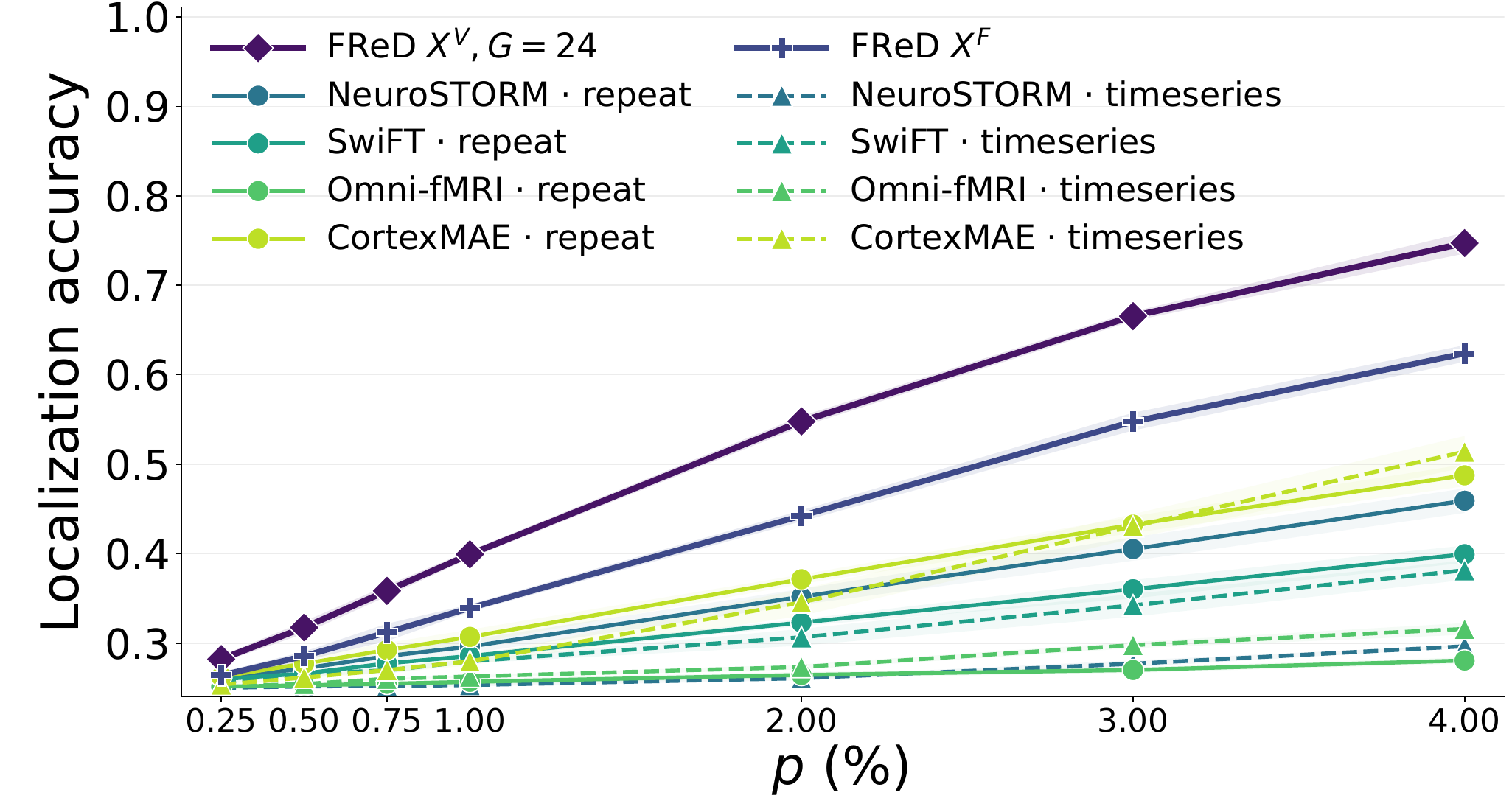}
        \\[1pt]{\small (b)} \vspace{-0.15in}
    \end{minipage}
\caption{
\textbf{(a)} Fraction of variance explained by the leading ImageNet principal directions, as a function of the fraction of components retained, shown separately for $\mathbf{X}^{V}$ (left) and $\mathbf{X}^{F}$ (right). Each panel includes an ImageNet reference computed on held-out ImageNet latents at the corresponding DCAE input resolution; a curve close to the ImageNet reference indicates a similar concentration of variance along the leading ImageNet principal directions. For $\mathbf{X}^{V}$, the curves are medians over the three anatomical slicing axes and the bands span their minimum and maximum.
\textbf{(b)} Four-way localization accuracy as a function of the injected percent signal change, averaged over the three spatial scales $\sigma$; chance accuracy is 25\%.
}
\label{fig:analysis}
\vspace{-0.2in}
\end{figure}

\section{Analyses}
\label{sec:analysis}
\subsection{Does the DCAE Latent Space Accommodate fMRI Variation?}
\label{subsec:pca_analysis}

We first examine how much fMRI variation is captured by the principal directions of DCAE representations learned from natural images.
We fit a PCA basis on DCAE latents from ImageNet test images and measure the fraction of fMRI variance explained by the leading $k$ ImageNet principal directions,
$\|\mathbf{H}\mathbf{V}_{1:k}\|_F^2 / \|\mathbf{H}\|_F^2$,
where $\mathbf{V}_{1:k}$ contains the leading $k$ ImageNet principal directions and $\mathbf{H}$ denotes the standardized latents of the domain being evaluated, centered by its own mean.
Since the DCAE latent size depends on the input resolution, we compare
each fMRI representation against an ImageNet reference computed at the
matching resolution: $3\times96\times96$ for volumetric slices and
$3\times224\times288$ for hemisphere maps.
The detailed setup can be found in \Cref{subsec:pca_details}.

As shown in \Cref{fig:analysis}(a), every fMRI representation falls below its ImageNet reference when only a small fraction of principal directions is retained, indicating that the directions explaining the largest variation in natural images account for a smaller fraction of fMRI variation.
The gap gradually narrows as more components are included, and by 60\% of the basis the retained ImageNet directions explain roughly $0.72$--$0.85$ of the variance across the fMRI representations.
Flat-map representations remain closer to their corresponding ImageNet reference than volumetric slices.
Overall, these results show that a substantial fraction of fMRI variation is recoverable from the ImageNet-derived basis, which supports using DCAE pre-trained with natural images to obtain fMRI representations.

\subsection{Are Localized Signal Changes Accessible from DCAE Representations?}
\label{subsec:gaussian_blob}
Downstream tasks may depend on small, spatially localized changes in the fMRI signal, and whether such changes survive a 32$\times$ spatial compression is not obvious.
We therefore inject controlled Gaussian perturbations of varying amplitude into resting-state HCP-A fMRI and evaluate whether their locations can be recovered from each frozen representation using a linear probe. At one of four fixed cortical centers, we add an isotropic Gaussian profile of width $\sigma$, restricted to the cortical mask, modulated by a canonical double-gamma hemodynamic response and scaled so that the added signal at the center equals a requested percent signal change $p$ of the local baseline. We sweep $p\in[0.25,4]\%$ and $\sigma\in\{2,3,4\}$. After injection we match the global statistics of each sequence to those of its clean counterpart, so that no probe can exploit a shift in intensity or variance. A four-way linear probe is then fitted on the frozen representations over four different splits and asked which center was perturbed. Each representation is given the frame at which the response peaks: {\ours} encodes it directly, while the foundation models, which require a temporal input, receive the same volume repeated across their window (\texttt{repeat}) or, as an additional condition, the full injected sequence (\texttt{timeseries}). The DCAE features are reduced to 768 dimensions by PCA, matching the largest foundation-model feature dimension. Full details are given in \Cref{subsec:blob_details}.

As shown in \Cref{fig:analysis}(b), all representations remain near chance at the weakest perturbation and improve monotonically as the injected signal becomes stronger. Both {\ours} variants stay above every foundation-model representation at every amplitude, with $\mathbf{X}^{V}$ the strongest throughout and $\mathbf{X}^{F}$ second; CortexMAE is the closest baseline, and neither input condition consistently outperforms the other across the foundation models. 
Information about localized signal changes therefore remains linearly accessible from the frozen DCAE features, even though the encoder has never been trained on fMRI. However, we note that this spatial-localization probe only measures the accessibility of perturbation, not the general representation quality.


\subsection{Task-dependent Normalization and Readout}
\label{subsec:task_dependency}










\paragraph{Normalization.}
\begin{wraptable}{r}{0.43\linewidth}
\vspace{-\intextsep}
\centering
\caption{
Linear probing results over 500 random splits on HBN-Movie/Sex.
}
\vspace{-0.15in}
\label{tab:hbn_movie_gender_lp}
\resizebox{\linewidth}{!}{%
\begin{tabular}{l|c|cc|cc}
\hline
\multirow{3}{*}{Method}
& \multirow{3}{*}{Input}
& \multicolumn{2}{c|}{\multirow{2}{*}{HBN-Movie}}
& \multicolumn{2}{c}{\multirow{2}{*}{HBN-Sex}} \\
& & & & & \\ \cline{3-6}
& & AUC $\uparrow$ & F1 $\uparrow$ & AUC $\uparrow$ & F1 $\uparrow$ \\ \hline
\oursl & $\mathbf{X}^V_g$
& \meanstd{.667}{.03}
& \meanstd{.614}{.04}
& \meanstd{.882}{.04}
& \meanstd{.741}{.06} \\
\oursl & $\mathbf{X}^V_{cf}$
& \meanstd{.863}{.03}
& \meanstd{.787}{.03}
& \meanstd{.822}{.04}
& \meanstd{.657}{.06} \\
\hline
\end{tabular}%
}
\vspace{-0.2in}
\end{wraptable}
Since HBN is a task-fMRI dataset, one may ask whether the sensitivity
to normalization observed in \Cref{subsec:state_prediction} stems from
the dataset itself rather than from the type of prediction task.
We therefore add a trait prediction task on the same dataset,
sex prediction of each subject, which let us compare the two task types under an identical data distribution.
As shown in Tab.~\ref{tab:hbn_movie_gender_lp}, HBN-Movie is far more sensitive to the choice of normalization than HBN-Sex; we assert the sensitivity therefore follows the prediction task rather than the dataset.
\paragraph{Readout.}
We showed that {\oursl} is ineffective for state prediction, which
motivated the temporal modeling in {\ourst}.
This raises the converse question of whether {\ourst} is better for
trait prediction.
\begin{wraptable}{r}{0.45\linewidth}
\vspace{-\intextsep}
\centering
\caption{
{\ourst} performance on trait prediction tasks.
\textbf{Bold} values indicate the best performance. LP: Linear probing, TFS: Train from scratch.
}
\label{tab:tablet_v2_on_resting}
\renewcommand{\arraystretch}{1.25}
\resizebox{\linewidth}{!}{%
\begin{tabular}{l|cc|cc|cc}
\hline
\multirow{3}{*}{Method}
& \multirow{3}{*}{Input}
& \multirow{3}{*}{Eval}
& \multicolumn{2}{c|}{ADHD-200}
& \multicolumn{2}{c}{HCP-A} \\ \cline{4-7}
& & & \multicolumn{2}{c|}{Diagnosis}
& \multicolumn{2}{c}{Intelligence} \\
& & & AUC $\uparrow$ & \multicolumn{1}{c|}{F1 $\uparrow$}
& MAE $\downarrow$ & $\rho$ $\uparrow$ \\ \hline
\oursl & $\mathbf{X}_{g}^V, \mathbf{X}_{cf}^V$ & \rowstrut LP
& \meanstd{.686}{.05}
& \meanstd{.604}{.05}
& \meanstd{\textbf{.633}}{.06}
& \meanstd{\textbf{.607}}{.05} \\
\oursl & $\mathbf{X}_{g}^F, \mathbf{X}_{cf}^F$ & \rowstrut LP
& \meanstd{\textbf{.717}}{.03}
& \meanstd{.607}{.05}
& \meanstd{.655}{.03}
& \meanstd{.576}{.04} \\
\hline
\ourst & $\mathbf{X}_g^V$ & TFS
& \meanstd{.663}{.05}
& \meanstd{.584}{.07}
& \meanstd{.652}{.03}
& \meanstd{.564}{.03} \\
\ourst & $\mathbf{X}_{cf}^V$ & TFS
& \meanstd{.637}{.05}
& \meanstd{.570}{.06}
& \meanstd{.664}{.06}
& \meanstd{.556}{.05} \\
\ourst & $\mathbf{X}_g^F$ & TFS
& \meanstd{.669}{.03}
& \meanstd{.609}{.05}
& \meanstd{.689}{.04}
& \meanstd{.474}{.08} \\
\ourst & $\mathbf{X}_{cf}^F$ & TFS
& \meanstd{.687}{.05}
& \meanstd{\textbf{.614}}{.09}
& \meanstd{.688}{.01}
& \meanstd{.505}{.03} \\
\hline
\end{tabular}%
}
\end{wraptable}

As shown in \Cref{tab:tablet_v2_on_resting}, {\ourst} does not improve over {\oursl} for trait prediction tasks.
Explicit temporal modeling therefore appears to matter specifically
where temporal dynamics are relevant to the target, whereas simple
temporal aggregation suffices for trait prediction.

\subsection{Component-wise Ablations}

\label{subsec:ablation_studies}

\begin{figure*}[!ht]
\centering
\begin{minipage}[b]{0.32\textwidth}
    \centering
    \includegraphics[width=\linewidth]{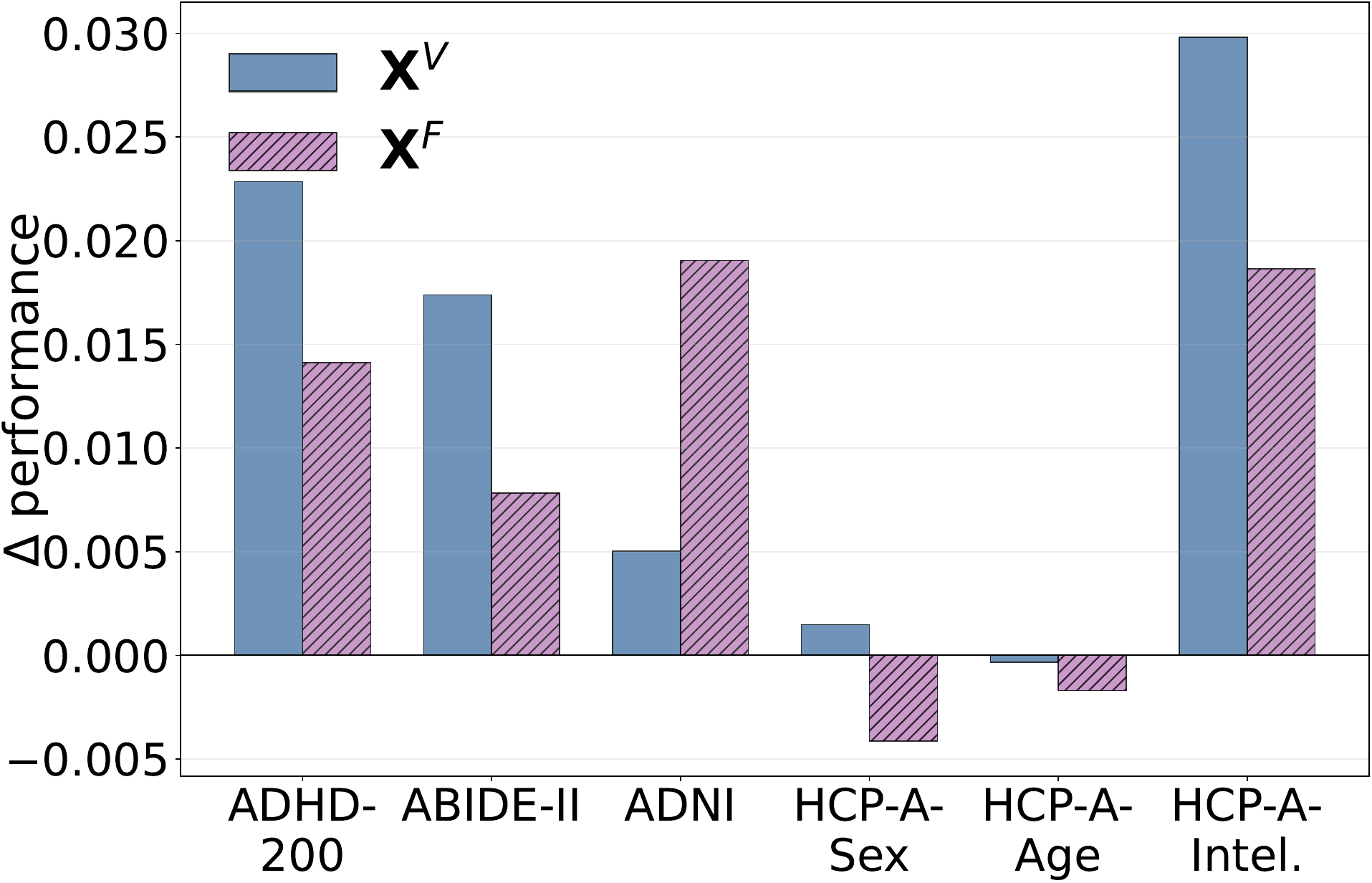}
    \\[2pt]{\small (a)}
\end{minipage}
\hfill
\begin{minipage}[b]{0.32\textwidth}
    \centering
    \includegraphics[width=\linewidth]{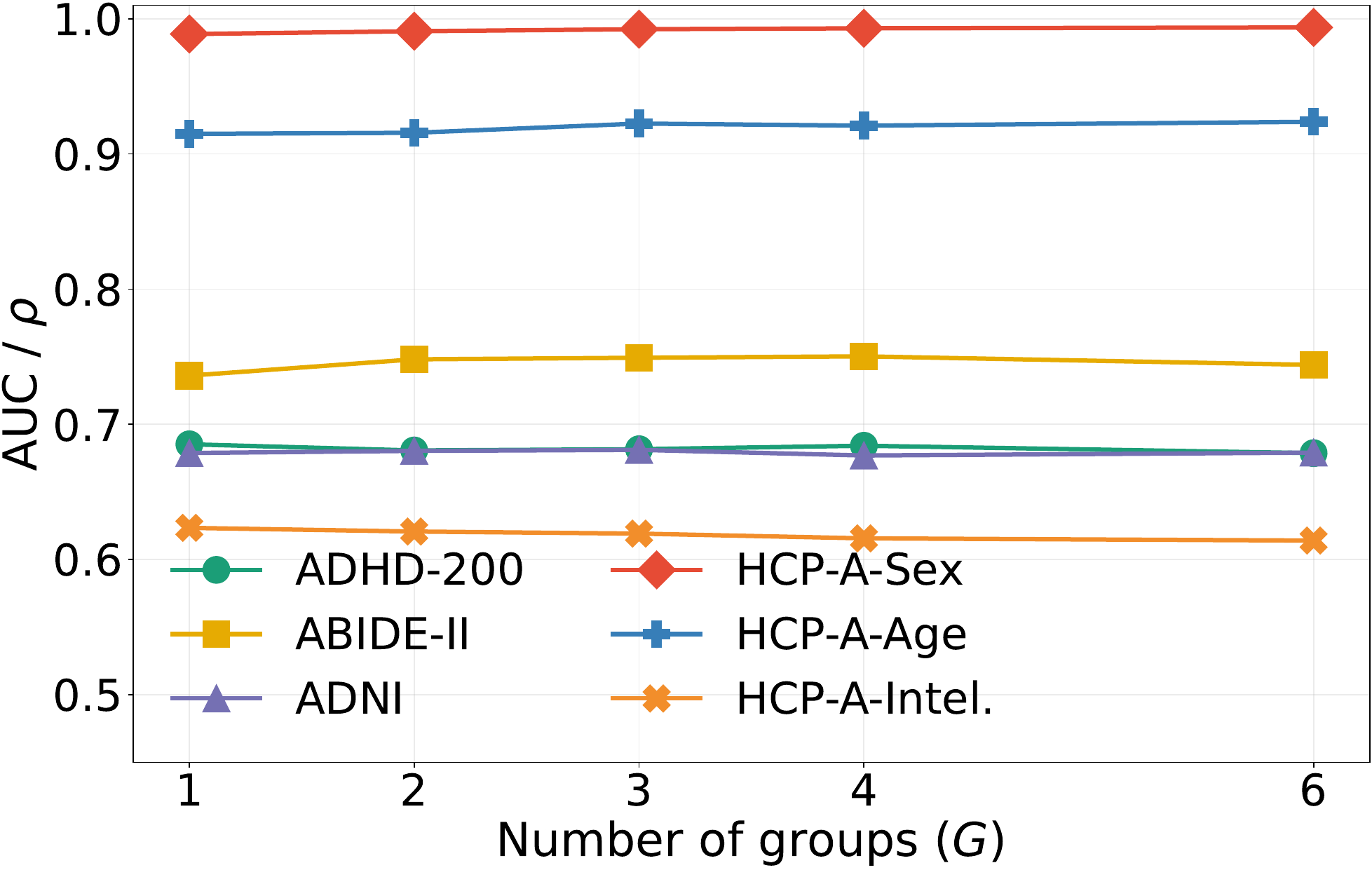}
    \\[2pt]{\small (b)}
\end{minipage}
\hfill
\begin{minipage}[b]{0.32\textwidth}
    \centering
    \includegraphics[width=\linewidth]{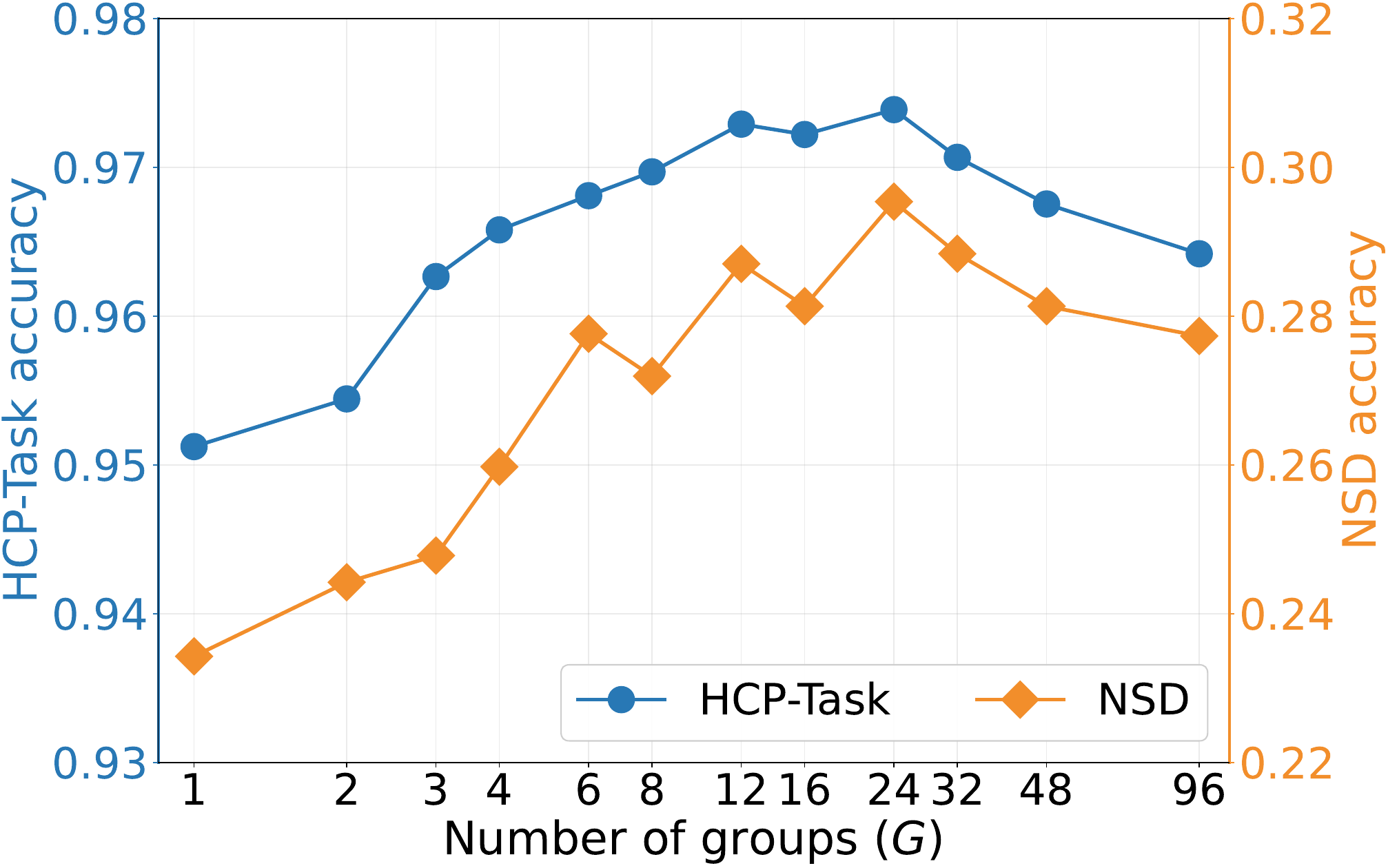}
    \\[2pt]{\small (c)}
\end{minipage}
\vspace{-0.1in}
\caption{
\textbf{(a)} Effect of the log-standard deviation feature on trait prediction, measured as the AUC or $\rho$ difference against replacing it with a zero vector.
\textbf{(b)} Trait prediction performance as a function of the number of groups $G$.
\textbf{(c)} State prediction accuracy as a function of $G$.
}
\label{fig:ablations}
\vspace{-0.2in}
\end{figure*}

\paragraph{Log-standard deviation features.}
We ablate the log-standard deviation feature by replacing it with a zero vector, which keeps the input dimensionality and parameter count unchanged.
As shown in \Cref{fig:ablations}(a), the log-standard deviation feature
generally improves performance, although the magnitude of the effect varies across tasks.

\paragraph{Number of groups $G$.}
We next vary the number of groups $G$, which controls how much spatial granularity along each slicing axis is retained in the frame representation.
Trait prediction is largely insensitive to $G$ and remains stable even with a single group per axis (\Cref{fig:ablations}(b)).
State prediction, in contrast, degrades substantially when $G$ is too small, but is stable across a moderate range
(\Cref{fig:ablations}(c)).

\paragraph{From TABLeT to \ourst.}
\begin{wraptable}{r}{0.4\linewidth}
\vspace{-\intextsep}
\centering
\caption{Comparison between TABLeT and {\ourst}, trained from scratch.}
\label{tab:tablet_v1_vs_v2}
\vspace{-0.1in}
\renewcommand{\arraystretch}{1.25}
\resizebox{\linewidth}{!}{%
\begin{tabular}{l|cc|c|c}
\hline
\multirow{3}{*}{Method}
& \multirow{3}{*}{Input}
& \multirow{3}{*}{Param.}
& \multicolumn{1}{c|}{\multirow{2}{*}{HCP-Task}}
& \multicolumn{1}{c}{\multirow{2}{*}{NSD}} \\
& & & & \\ \cline{4-5}
& & & ACC & ACC \\ \hline
TABLeT
& $\mathbf{X}^V_{cf}$
& 129.5M
& \meanstd{94.2}{0.2}
& \meanstd{23.9}{1.0} \\ \hline
TABLeT with $\mathbf{\bar Z}^V_{a,t}$
& $\mathbf{X}^V_{cf}$
& 145.4M
& \meanstd{94.7}{0.3}
& \meanstd{27.5}{0.4}\\
\ourst
& $\mathbf{X}^V_{cf}$
& 78.2M
& \meanstd{\textbf{97.4}}{0.1}
& \meanstd{\textbf{29.5}}{0.2} \\
\hline
\end{tabular}%
}
\vspace{-0.2in}
\end{wraptable}
Finally, we identify where the improvement of {\ourst} over TABLeT
comes from.
Since TABLeT operates on volumetric fMRI, we conduct this ablation with
$\mathbf{X}^{V}$.
The two models differ in their tokenization and in architectural and regularization choices.
To isolate the former, we replace TABLeT's tokenization with ours,
constructing a single token per frame from group-averaged DCAE latents,
while keeping its temporal Transformer.
As reported in \Cref{tab:tablet_v1_vs_v2}, this intermediate model already improves performance over the
original TABLeT. Moving from this intermediate model to FReD-State yields further gains
while reducing the parameter count.
Because this second step introduces several architectural and regularization changes simultaneously, we attribute the additional gain
to these remaining changes jointly rather than to any single component,  detailed in \Cref{sec:tablet_arch}.



\section{Conclusion and Limitations}
\label{sec:conclusion}
We asked how much fMRI-specific representation pre-training is necessary for strong downstream prediction. We studied this question using a frozen DCAE pre-trained exclusively on natural images with task-matched readouts. For trait prediction, temporal mean and log-standard deviation followed by a linear predictor achieve performance competitive with fully fine-tuned fMRI foundation models. For state prediction, a shallow Transformer over one token per frame performs comparably to the strongest foundation models, while using a smaller readout and a shorter attention sequence than TABLeT. These results show that strong performance on current benchmarks can be achieved without fMRI-specific representation pre-training, and establish frozen natural-image representations as a baseline against which the benefits of large-scale fMRI pre-training should be measured.

However, several questions remain open.
The Gaussian injection analysis is synthetic and spatially localized,
so it speaks to linear accessibility rather than to downstream ranking.
Relatedly, we do not resolve when each data type should be preferred,
although we observe that $\mathbf{X}^{F}$ tends to be stronger on state
prediction whereas $\mathbf{X}^{V}$ recovers injected perturbations far
more accurately and leads on the HCP-A phenotype tasks.
We also treat the encoder as fixed throughout, and do not explore whether adapting it to fMRI could further improve performance.
Despite this, our results show that a frozen natural-image autoencoder can produce strong representations for fMRI, and suggest that the benefit of fMRI-specific pre-training should be demonstrated against such baselines rather than assumed.

\subsection*{AI use Statement}
In this work, we used generative AI tools for providing feedback on research methodology or experiments, implementing methods, and assisting with translation.
We have not used generative AI tools for proposing or refining hypotheses, supporting qualitative or thematic data analysis, and interpreting results.
Generating synthetic data sets, helping develop theoretical models or conceptual frameworks, formulating mathematical claims, providing critical ingredients for proving mathematical claims, assisting in the writing of proofs, and cleaning and reformat dataset are not applicable to this work.

Additionally, we used generative AI tools for modifying scientific figures or images, creating or editing software code, drafting parts of a research paper, editing a research paper to improve readability. All AI-assisted outputs were carefully reviewed by the authors. In particular, 
the authors created the initial versions of all scientific figures and images themselves and used LLM-based tools only to improve their visual clarity and presentation.
In addition, the authors independently verified LLM-generated code and tested it for correctness. The authors also independently reviewed all LLM-generated drafts, using them only as starting points before revising and finalizing the text themselves. and write a final script by own. We take full responsibility for the final content of this work, including all text, claims, code, figures, and other artifacts produced with the assistance of generative AI.

\subsection*{Ethics Statement}
This work relies exclusively on publicly available, open-source fMRI datasets. We did not collect any new human-subject data or interact directly with study participants. Therefore, no additional ethical approval was required for this study.

\subsection*{Reproducibility Statement}
We provide detailed experimental settings in \Cref{sec:experimental_details}.
We plan to release our code, preprocessing scripts, and data splits after the review process, subject to the licenses and access policies
of the original datasets.

\bibliography{iclr2027_conference}
\bibliographystyle{iclr2027_conference}

\appendix
\clearpage
\renewcommand\thesection{\Alph{section}}
\setcounter{section}{0}
{
\centering
\Large
\textbf{Natural Image Autoencoder-Based fMRI Representations for Trait and State Prediction}\\
\vspace{0.5em}Appendix \\
\vspace{1.0em}
}

\section{Related Work}
\subsection{Utilizing Natural Image Prior for Medical Imaging}
The limited availability and diversity of medical imaging data have long posed challenges for learning generalizable representations directly from the medical domain. In contrast, foundation models pre-trained on large-scale natural image datasets benefit from substantially richer and more diverse visual supervision, yielding robust representations. Motivated by this discrepancy, recent studies have investigated transferring such pre-trained representations to medical imaging, rather than training modality-specific foundation models solely on medical data.

Natural-image priors have proven effective even for volumetric medical data despite considerable domain and dimensionality gaps. Raptor \citep{raptor} leverages a frozen 2D foundation model pre-trained on natural images to extract representations from orthogonal slices of 3D medical volumes and compresses the resulting features through random projections. This training-free approach demonstrates that representations learned from natural images can effectively support diverse downstream tasks on 3D medical data. AnyMC3D \citep{anymc3d} further extends this direction to 3D medical image classification, adapting a pre-trained 2D foundation model using lightweight task-specific LoRA modules and attention-based slice aggregation. Notably, it shows that properly adapted general-purpose models can match or even outperform medical-specific foundation models across diverse 3D classification tasks. More recently, TABLeT \citep{tablet, tabletW} employs a pre-trained 2D natural-image autoencoder as an off-the-shelf tokenizer for fMRI volumes, transforming high-dimensional 3D brain images into compact tokens for modeling long-range temporal dynamics. Together, these studies highlight the potential of transferring rich visual priors learned from natural images to medical imaging, offering an alternative to costly domain-specific pre-training under limited medical data.

\section{Implementation Details}
\label{sec:experimental_details}

\subsection{Datasets}
\label{subsec:datasets_detail}
\begin{table*}[ht]
    \centering
    \caption{
        Summary of the datasets used for trait prediction (top) and
        state prediction (bottom). The numbers of subjects and samples
        represent the totals after preprocessing. Duration indicates
        the temporal duration of each sample. Trait prediction datasets
        contain one sample per subject, whereas state prediction
        datasets contain multiple samples per subject. Dx denotes
        diagnosis classification; Task21 denotes 21-class cognitive
        task-state decoding; and COCO24 denotes 24-class object-category
        decoding. The majority-class proportion is omitted for
        regression tasks.
    }
    \label{tab:dataset_desc}
    \begin{tabular}{llccccc}
        \hline
        Dataset & Target & Subjects & Samples & Duration
        & \# Classes & Majority (\%) \\
        \hline
        ADHD-200 & ADHD Dx
        & 430 & 430 & 300 s & 2 & 56.3 \\
        ABIDE-II & ASD Dx
        & 616 & 616 & 300 s & 2 & 56.3 \\
        ADNI & MCI Dx
        & 464 & 464 & 300 s & 2 & 59.5 \\
        HCP-A & Sex
        & 1,069 & 1,069 & 382.4 s & 2 & 57.1 \\
        HCP-A & Age, Intelligence
        & 1,069 & 1,069 & 382.4 s & -- & -- \\
        \hline
        HBN-Movie & Movie
        & 428 & 856 & 200 s & 2 & 50.0 \\
        HCP-Task & Task21
        & 604 & 26,995 & 16 s & 21 & 16.9 \\
        NSD & COCO24
        & 8 & 47,589 & 16 s & 24 & 6.8 \\
        \hline
    \end{tabular}
\end{table*}
The detailed description of each dataset is provided in \Cref{tab:dataset_desc}.

\paragraph{Preprocessing.}
For ADHD-200~\citep{ADHD}, ABIDE-II~\citep{ABIDE2}, ADNI~\citep{ADNI},
and HBN~\citep{HBN}, we perform minimal preprocessing using
fMRIPrep~\citep{fmriprep1, fmriprep2}.
For ADHD-200, ABIDE-II, and ADNI, we additionally regress out nuisance
signals using cosine basis functions, six rigid-body motion parameters,
and anatomical CompCor components.
For HCP-A, we use the minimally processed data provided
by~\citet{HCPA}, and for HCP-Task and NSD, the preprocessed data
provided by BrainMarks~\citep{cortexmae}.
We then resample all fMRI data to match the TR each foundation model
was pre-trained with.

\paragraph{Data splits.}
For ADHD-200, ABIDE-II, and ADNI, we perform stratified sampling to
preserve the distributions of acquisition sites and diagnostic labels
across the splits, and for HCP-A we preserve the distributions of sex,
age, and intelligence.
HBN is split randomly, whereas for HCP-Task and NSD we use the fixed
splits provided by BrainMarks~\citep{cortexmae}.
All splits are constructed at the subject level, so that no subject
appears in more than one split.
For the linear-probing experiments in \Cref{fig:lp_results}, \Cref{tab:hbn_movie_gender_lp}, and \Cref{fig:lp_rp768_results}, we evaluate each method over 500
independently sampled train--test splits, assigning 85\% of the data to
training and the remaining 15\% to testing.
All other experiments use a 70/15/15 train/validation/test ratio,
repeated over 12 splits for the clinical diagnosis tasks and four
splits for HCP-A and HBN-Movie; HCP-Task and NSD instead use three
random seeds on their fixed splits. 

\subsection{Training and Evaluation}
\label{subsec:train_eval_details}

\paragraph{DCAE.}
We use the unmodified \texttt{dc-ae-f32c32-in-1.0} checkpoint provided
by~\citet{dcae} in all experiments, and keep it frozen throughout.

\paragraph{Windowing.}
For the $\mathbf{X}^V$- and $\mathbf{X}^F$-based models, training uses
randomly sampled consecutive frames.
At evaluation time, following~\citet{swift, tablet}, we average the
model outputs (logits) over all windows starting from the first frame.
For linear probing, we follow the protocol of
BrainMarks~\citep{cortexmae}, which uses the window-averaged embedding.
If inputs are shorter than a model’s expected temporal sequence length, we pad the input with the per-coordinate mean, following \citet{cortexmae}.

\paragraph{Late fusion.}
We select $\alpha$ without using the test set at any stage. For experiments with an 85/15 train/test split, $\alpha$ is selected using out-of-fold predictions obtained from five-fold cross-validation on the training set. For experiments with a 70/15/15 train/validation/test split, it is selected on the validation set.

Because the search is one-dimensional, we first evaluate the selection criterion on a uniform grid over $\alpha \in [0,1]$ with a step size of $0.1$. For smooth criteria, such as mean absolute error and weighted cross-entropy, the best-performing grid point is used to initialize an SLSQP \citep{SLSQP} refinement, yielding a continuous value of $\alpha$. For rank-based criteria such as AUROC, we use the best grid point directly.

\paragraph{Shared settings.}
For neural-network training, unless stated otherwise, all experiments use the following settings.
\begin{compactitem}
\item \texttt{Target normalization}: For regression tasks, we z-normalize the target values using the mean and standard deviation computed from the training split.

\item \texttt{Optimizer}: AdamW with a cosine decay learning rate
      schedule and weight decay $10^{-2}$.
      We use \texttt{fp16} mixed precision,
      \texttt{BCEWithLogitsLoss} for binary classification, and
      \texttt{L1Loss} for regression.
      For ADHD-200, ABIDE-II, and ADNI, we enable the \texttt{pos-weight}
      option to account for class imbalance.
      For HCP-Task and NSD, we use \texttt{CrossEntropyLoss}.

\item \texttt{Hyperparameter search}: We select hyperparameters per
      model on the validation set, using validation loss for ADHD-200,
      ABIDE-II, and ADNI (so that \texttt{pos-weight} is reflected),
      AUROC for HCP-A-Sex and HBN-Movie, accuracy for HCP-Task and NSD,
      and MAE for HCP-A-Age and HCP-A-Intelligence.
      For every model and dataset, we search at least eight learning
      rates on a log-uniform grid within the range listed below.

\item \texttt{Early stopping}: We report the early-stopped model for
      all methods except SwiFT, for which we report the final epoch,
      as we observed its training to be stable.
\end{compactitem}

\paragraph{Brain-JEPA}
\begin{compactitem}
    \item \texttt{Learning rate}: Selected from $[1\times10^{-6},\,3\times10^{-4}]$.
    \item \texttt{Batch size}: 16.
    \item \texttt{Epochs}: 50.
\end{compactitem}

\paragraph{BrainHarmonix-F}
\begin{compactitem}
    \item \texttt{Learning rate}: Selected from $[3\times10^{-8},\,5\times10^{-3}]$.
    \item \texttt{Batch size}: 8.
    \item \texttt{Epochs}: 50.
\end{compactitem}

\paragraph{Brain-DiT}
\begin{compactitem}
    \item \texttt{Learning rate}: Selected from $[1\times10^{-6},\,3\times10^{-4}]$.
    \item \texttt{Batch size}: 2.
    \item \texttt{Epochs}: 20.
\end{compactitem}

\paragraph{SwiFT}
\begin{compactitem}
    \item \texttt{Learning rate}: Selected from $[3\times10^{-8},\,9\times10^{-5}]$.
    \item \texttt{Batch size}: 8.
    \item \texttt{Epochs}: 30.
    \item \texttt{Window size}: 20.
    \item \texttt{TR}: 0.72s
\end{compactitem}

\paragraph{NeuroSTORM}
\begin{compactitem}
    \item \texttt{Learning rate}: Selected from $[3\times10^{-8},\,9\times10^{-5}]$.
    \item \texttt{Batch size}: 8.
    \item \texttt{Epochs}: 30.
    \item \texttt{Window size}: 20.
    \item \texttt{TR}: 0.72s
\end{compactitem}

\paragraph{Omni-fMRI}
\begin{compactitem}
    \item \texttt{Learning rate}: Selected from
    $[3\times10^{-8},\,9\times10^{-5}]$ for trait prediction,
    $[3\times10^{-8},\,5\times10^{-4}]$ for HBN-Movie, and
    $[3\times10^{-7},\,5\times10^{-4}]$ for HCP-Task and NSD.
    The learning rate of the prediction head is set to $10\times$ the
    base learning rate.
    \item \texttt{Batch size}: 8 for trait prediction and HBN-Movie,
    and 16 for HCP-Task and NSD.
    \item \texttt{Epochs}: 30.
    \item \texttt{Window size}: 40.
    \item \texttt{TR}: 0.8s
\end{compactitem}

\paragraph{CortexMAE}
\begin{compactitem}
    \item \texttt{Learning rate}: Selected from
    $[3\times10^{-8},\,9\times10^{-5}]$ for trait prediction,
    $[3\times10^{-8},\,9\times10^{-6}]$ for HBN-Movie, and
    $[3\times10^{-8},\,9\times10^{-5}]$ for HCP-Task and NSD.
    \item \texttt{Batch size}: 8 for trait prediction and HBN-Movie,
    and 64 for HCP-Task and NSD.
    \item \texttt{Epochs}: 30.
    \item \texttt{Window size}: 16.
    \item \texttt{TR}: 1s.
\end{compactitem}

\paragraph{TABLeT}
\begin{compactitem}
    \item \texttt{Learning rate}: Selected from
    $[5\times10^{-9},\,5\times10^{-6}]$ for trait prediction and
    HBN-Movie, and $[7\times10^{-6},\,2\times10^{-4}]$ for HCP-Task and
    NSD.
    \item \texttt{Batch size}: 4 for trait prediction and HBN-Movie,
    and 16 for HCP-Task and NSD.
    \item \texttt{Epochs}: 30 for trait prediction and HBN-Movie, 50
    for HCP-Task, and 15 for NSD.
    \item \texttt{Window size}: We use full frames.
    \item \texttt{TR}: 0.8s for trait prediction and HBN, 1s for HCP-Task and NSD.
\end{compactitem}

\paragraph{\oursl}
\begin{compactitem}
    \item \texttt{Hyperparameters}: We use scikit-learn's default search space.
    \item \texttt{TR} : 0.8s for trait prediction and HBN.
\end{compactitem}

\paragraph{\ourst}
\begin{compactitem}
    \item \texttt{Learning rate}: Selected from
    $[1\times10^{-6},\,1\times10^{-4}]$ for HBN-Movie, and
    $[7\times10^{-6},\,2\times10^{-4}]$ for HCP-Task and NSD. For the trait prediction tasks in \Cref{subsec:task_dependency}, the learning rate is selected from $[5\times10^{-9},\,5\times10^{-6}]$ matched with TABLeT's search space.
    \item \texttt{Batch size}: 8 for trait prediction and HBN-Movie,
    and 16 for HCP-Task and NSD.
    \item \texttt{Epochs}: 50 for trait prediction, HBN-Movie, and
    HCP-Task, and 15 for NSD.
    \item \texttt{Window size} : We use full frames.
    \item \texttt{TR} : 0.8s for trait prediction and HBN, 1s for HCP-Task and NSD.
\end{compactitem}
\subsection{PCA Analysis Details}
\label{subsec:pca_details}
We extract DCAE latents from 100k ImageNet test images and from HCP-A,
using only the first temporal frame of each fMRI sequence to keep the
computation tractable.
For HCP-A we consider $\mathbf{X}^{r}_{n}$ with $r\in\{V,F\}$ and
$n\in\{g,cf\}$: for $\mathbf{X}^{V}_{n}$ we take the center sagittal,
coronal, and axial slices, evaluate each axis separately, and report
their median and range in \Cref{fig:analysis}(a), and for
$\mathbf{X}^{F}_{n}$ we treat the two hemispheres as independent
samples.
The ImageNet latents are randomly split into 80k samples for
standardization and PCA fitting and 20k held-out samples for
evaluation.


\subsection{Gaussian Injection Experiments}
\label{subsec:blob_details}
The experiment uses all HCP-A subjects and a 16\,s analysis
window sampled at $\mathrm{TR}=0.8$\,s, giving $T=20$ frames.
Since the procedure is identical for every subject, we describe it for
a single subject and omit the subject index.

\paragraph{Injection centers.}
The four centers are shared across every model, $p,\sigma$, and data split.
They are selected from the intersection of the cohort-wide anatomical
support and a cortical atlas mask $C$, subject to a minimum distance of
8 voxels from the outer support boundary and a minimum pairwise
separation of 12 voxels.
In native LAS-oriented MNI152 2-mm voxel indices they are
$\mathbf{c}_1=(60,39,34)$, $\mathbf{c}_2=(20,47,33)$,
$\mathbf{c}_3=(31,30,34)$, and $\mathbf{c}_4=(24,38,43)$, with a
realized minimum pairwise distance of 13.93 voxels.

\paragraph{Signal injection.}
Let $\mathbf{X}_t(\mathbf{v})$ denote the clean BOLD value at voxel
$\mathbf{v}$ in frame $t$, in the native MNI volume and before any
normalization.
For center $\mathbf{c}_j$, spatial scale $\sigma$, and percent signal change $p$, the injected frame is
\begin{equation}
  \mathbf{X}^{\mathrm{inj}}_t(\mathbf{v})
  =
  \mathbf{X}_t(\mathbf{v})
  +
  b_j\tfrac{p}{100}\,
  h(t)\,
  A_{j,\sigma}(\mathbf{v}),
  \label{eq:gaussian_psc_injection}
\end{equation}
which adds a fixed spatial profile $A_{j,\sigma}$, modulated by a fixed
time course $h$ and scaled by an amplitude $b_j\cdot p/100$.
The clean condition leaves $\mathbf{X}_t$ unchanged.
The three factors are defined below.

\paragraph{Spatial profile.}
$A_{j,\sigma}$ is an isotropic Gaussian centered at $\mathbf{c}_j$,
restricted to the cortical atlas mask $C$ and to the subject's
anatomical support (foreground) $M$:
\begin{equation}
  A_{j,\sigma}(\mathbf{v})
  =
  \exp\!\left(
    -\frac{\lVert \mathbf{v}-\mathbf{c}_j\rVert_2^2}{2\sigma^2}
  \right)
  C(\mathbf{v})\,M(\mathbf{v}).
  \label{eq:gaussian_spatial_profile}
\end{equation}

The perturbation is thus confined to cortex: both the centers and the
Gaussian tails are restricted to $C$.

\paragraph{Amplitude and the definition of PSC.}
$b_j$ is the local baseline intensity at $\mathbf{c}_j$, taken as the
mean of the temporal-mean image
$\overline{\mathbf{X}}(\mathbf{v})=\tfrac{1}{T}\sum_{t=1}^{T}
\mathbf{X}_t(\mathbf{v})$ over the nonzero supported voxels in the
$7\times7\times7$ neighborhood $\mathcal{N}_j$ centered on it,
$b_j=|\mathcal{N}_j|^{-1}\sum_{\mathbf{v}\in\mathcal{N}_j}
\overline{\mathbf{X}}(\mathbf{v})$.
Since $A_{j,\sigma}(\mathbf{c}_j)=1$ and $\max_t h(t)=1$, the added
signal at the Gaussian center and at the frame where the response peaks
is exactly $p$ percent of $b_j$, before the global-statistics matching
described below.

\paragraph{Time course.}
$h(t)$ is a canonical SPM hemodynamic response convolved with a 4\,s boxcar $1_{[0,4)}$ and normalized to unit peak,
\begin{equation}
  h(t) \propto {1}_{[0,4)} *
  \left(
    \operatorname{GammaPDF}(\cdot\,;6,1)
    - \tfrac{1}{6}\operatorname{GammaPDF}(\cdot\,;16,1)
  \right),
  \qquad \max_t h(t)=1.
\end{equation}

\paragraph{Removing global-intensity shortcuts.}
After injection we match the mean and standard deviation of the
modified sequence to those of its clean counterpart, computed jointly
over all frames and all voxels in $M$:
\begin{equation}
  \widehat{\mathbf{X}}^{\mathrm{inj}}_t(\mathbf{v})
  =
  \frac{\mathbf{X}^{\mathrm{inj}}_t(\mathbf{v})-\mu_{\mathrm{inj}}}
       {\sigma_{\mathrm{inj}}+\epsilon}
  \left(\sigma_{\mathrm{clean}}+\epsilon\right)
  +
  \mu_{\mathrm{clean}},
  \qquad \mathbf{v}\in M,
  \label{eq:gaussian_global_stat_matching}
\end{equation}
with values outside $M$ reset to zero afterwards.
The matched sequence is then coordinate--frame normalized, yielding $\mathbf{X}^{V}_{cf}$.

\paragraph{Frozen representations.}
For NeuroSTORM, SwiFT, Omni-fMRI, and CortexMAE we extract two frozen
representations.
The \texttt{repeat} condition fills the model's temporal input with the
frame at which $h(t)$ peaks, removing genuine temporal evolution while
preserving the peak spatial perturbation, whereas the \texttt{timeseries}
condition adapts the full injected sequence to each checkpoint's
temporal sampling and input length.

The two {\ours} variants use the peak frame only, since the DCAE
encodes frames independently and the repeat condition is its natural
counterpart.
For $\mathbf{X}^{V}_{cf}$ we compute $\mathbf{Z}^{V}_{cf,t}$ following
\Cref{eq:volume_frame_latent} with $G=24$, and for
$\mathbf{X}^{F}_{cf}$ we project the matched volume to the fsLR surface
and compute $\mathbf{Z}^{F}_{cf,t}$ following
\Cref{eq:flat_map_frame_latent}, in both cases at the peak frame.
Both are reduced to 768 dimensions by PCA fitted on the training
subjects only, matching the largest foundation-model feature dimension.

\paragraph{Probing protocol.}
Each subject contributes one injected sample at each of the four
centers, giving 4,276 balanced samples scored by accuracy against a
chance level of 25\%.
We use four predefined subject-disjoint 70/15/15 train/validation/test splits and fit a
\texttt{LogisticRegression} probe from scikit-learn.

\section{Additional Experimental Results}
\label{sec:additional_results}

\subsection{Dimension-reduced Linear Probing}
\label{subsec:dim_matched_lp}
\begin{figure}[t]
    \centering
    \includegraphics[width=\linewidth]{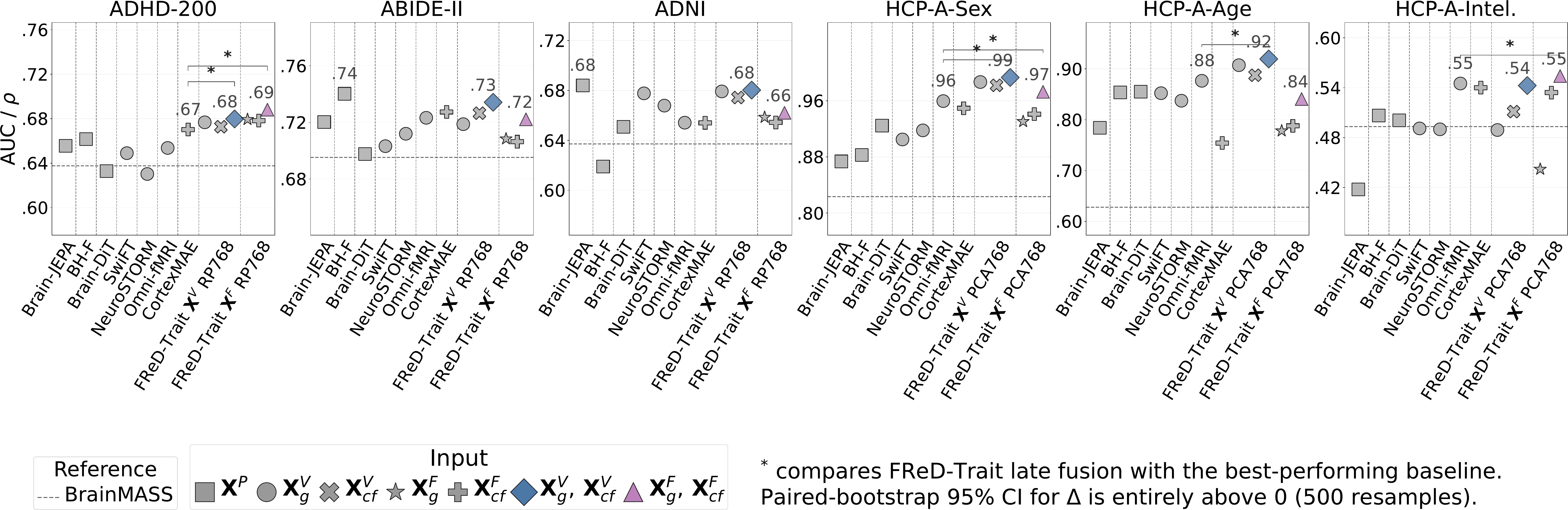}
\caption{
Linear-probing results over 500 random train-test splits after
reducing the {\oursl} features to 768 dimensions, reported as AUC for
classification and $\rho$ for regression.
RP768 and PCA768 denote Gaussian random projection and PCA,
respectively.
}
    \label{fig:lp_rp768_results}
    \vspace{-0.25in}
\end{figure}
The frame representations of {\oursl} are higher-dimensional than those
of the foundation models, which could by itself account for its
linear-probing performance in \Cref{subsec:trait_prediction}.
To assess this effect, we reduce the {\oursl} features to 768 dimensions,
matching the dimensionality of CortexMAE, and repeat the same
500-split protocol.
The reduction method depends on the sample size: with 85\% of the data
used for training, the clinical datasets provide only 365--523 training
samples, so a 768-component PCA is not defined and we use Gaussian
random projection instead.
For HCP-A, where 908 training samples are available, we use PCA.

As shown in Fig.~\ref{fig:lp_rp768_results}, the competitiveness of
FReD-Trait is largely preserved at matched dimensionality.
On ADHD-200, HCP-A sex, and HCP-A age, the reduced representations remain significantly better than the strongest baseline.
Performance does drop on ABIDE-II and HCP-A intelligence,
where the reduced features fall to the level of the best baselines,
indicating that part of the margin on these tasks may come from the
larger feature dimension.
Overall, however, FReD-Trait remains competitive with fully fine-tuned
foundation models after reducing its features to the largest foundation-model feature dimension, so its linear-probing performance cannot be attributed to dimensionality alone.

\subsection{Late Fusion Weight Distribution}
\label{subsec:fusion_weights}
\begin{wrapfigure}{r}{0.5\linewidth}
\vspace{-\intextsep}
\centering
\includegraphics[width=\linewidth]
{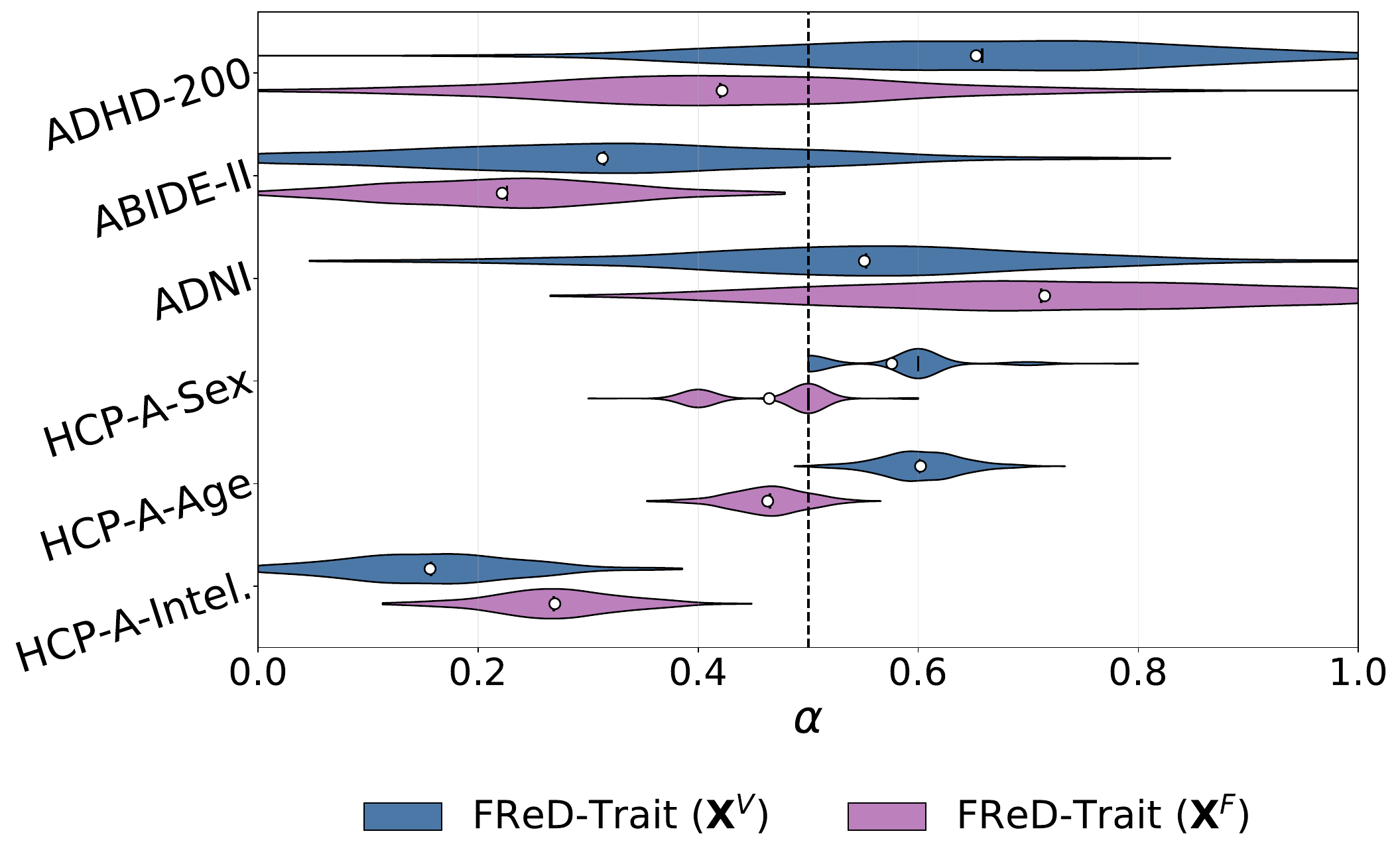}
\caption{
Distribution of the late-fusion weight $\alpha$ over the 500
train-test splits, shown separately for {\oursl} with
$\mathbf{X}^{V}$ and $\mathbf{X}^{F}$.
Larger $\alpha$ places more weight on the globally normalized branch,
and the dashed line marks equal weighting.
White markers indicate the mean.
}
\vspace{-0.2in}
\label{fig:late_fusion_weights}
\end{wrapfigure}
We analyze the fusion weight $\alpha$ selected for each task, where
larger $\alpha$ favors the globally normalized branch and smaller
$\alpha$ the coordinate--frame-normalized one.
As shown in \Cref{fig:late_fusion_weights}, the selected weights vary
substantially across tasks.
ASD classification and intelligence prediction consistently favor
coordinate--frame normalization under both data types, whereas ADHD and
ADNI diagnosis and the HCP-A sex and age tasks place at least as much weight on the globally normalized branch.

The distributions are broad for the clinical diagnosis tasks and much
narrower for the HCP-A tasks, mirroring their split-to-split
variability.
Across all tasks, however, the mean weight rarely approaches $0$ or
$1$, suggesting that both normalization schemes can contribute
complementary information.
These results highlight normalization as an important component of
the downstream pipeline: the two schemes retain complementary
information, and their relative utility varies across tasks.

\subsection{Partial Fine-tuning}
\label{subsec:partial_ft}
\definecolor{lprow}{HTML}{FFF6CC}
\begin{table*}[!ht]
\caption{
Partial fine-tuning results on trait prediction tasks. Each foundation model is partially fine-tuned, while {\oursl} uses linear probing on frozen features. Results are reported as the mean and standard deviation across 12 data splits for the clinical diagnosis tasks and four splits for HCP-A. \textbf{Bold} and \underline{underlined} values indicate the best and second-best performance for each metric.
}
\centering
\setlength{\tabcolsep}{3pt}
\renewcommand{\arraystretch}{1.25}
\resizebox{\linewidth}{!}{
\begin{tabular}{l|c|cc|cc|cc|cc|cc|cc}
\hline
\multirow{3}{*}{Method} & \multirow{3}{*}{Input}
 & \multicolumn{2}{c|}{ADHD-200} & \multicolumn{2}{c|}{ABIDE-II} & \multicolumn{2}{c|}{ADNI}
 & \multicolumn{6}{c}{HCP-A} \\ \cline{3-14}
 & & \multicolumn{2}{c|}{Diagnosis} & \multicolumn{2}{c|}{Diagnosis} & \multicolumn{2}{c|}{Diagnosis}
 & \multicolumn{2}{c|}{Sex} & \multicolumn{2}{c|}{Age} & \multicolumn{2}{c}{Intelligence} \\
 & & AUC $\uparrow$ & \multicolumn{1}{c|}{F1 $\uparrow$} & AUC $\uparrow$ & \multicolumn{1}{c|}{F1 $\uparrow$} & AUC $\uparrow$ & \multicolumn{1}{c|}{F1 $\uparrow$}
 & AUC $\uparrow$ & \multicolumn{1}{c|}{F1 $\uparrow$} & MAE $\downarrow$ & \multicolumn{1}{c|}{$\rho$ $\uparrow$} & MAE $\downarrow$ & $\rho$ $\uparrow$ \\ \hline

\rowcolor{white}Brain-JEPA & $\mathbf{X}^P$ \rowstrut
 & \meanstd{.662}{.04} & \meanstd{.595}{.03} & \meanstd{.721}{.03} & \meanstd{.572}{.06} & \meanstd{.674}{.04} & \meanstd{.527}{.07}
 & \meanstd{.905}{.02} & \meanstd{.817}{.01} & \meanstd{.454}{.01} & \meanstd{.830}{.01} & \meanstd{.704}{.04} & \meanstd{.429}{.05} \\ \hline

\rowcolor{white}BH-F & $\mathbf{X}^P$ \rowstrut
 & \meanstd{.662}{.05} & \meanstd{\underline{.617}}{.03} & \meanstd{.710}{.03} & \meanstd{.593}{.05} & \meanstd{.614}{.06} & \meanstd{.421}{.10}
 & \meanstd{.896}{.01} & \meanstd{.781}{.01} & \meanstd{.430}{.02} & \meanstd{.839}{.01} & \meanstd{.677}{.04} & \meanstd{.502}{.05} \\ \hline

\rowcolor{white}Brain-DiT & $\mathbf{X}^P$ \rowstrut
 & \meanstd{.650}{.03} & \meanstd{.604}{.04} & \meanstd{.704}{.04} & \meanstd{.558}{.09} & \meanstd{.668}{.05} & \meanstd{.562}{.04}
 & \meanstd{.936}{.03} & \meanstd{.856}{.01} & \meanstd{.408}{.02} & \meanstd{.863}{.01} & \meanstd{.674}{.02} & \meanstd{.517}{.02} \\ \hline

\rowcolor{white}SwiFT & $\mathbf{X}^V_g$ \rowstrut
 & \meanstd{.654}{.06} & \meanstd{\textbf{.618}}{.04} & \meanstd{.705}{.03} & \meanstd{.588}{.04} & \meanstd{\textbf{.699}}{.07} & \meanstd{\textbf{.613}}{.08}
 & \meanstd{.945}{.02} & \meanstd{.839}{.02} & \meanstd{\underline{.391}}{.02} & \meanstd{\underline{.866}}{.01} & \meanstd{.683}{.03} & \meanstd{.500}{.03} \\ \hline

\rowcolor{white}NeuroSTORM & $\mathbf{X}^V_g$ \rowstrut
 & \meanstd{.647}{.04} & \meanstd{.583}{.06} & \meanstd{.706}{.03} & \meanstd{.599}{.04} & \meanstd{.652}{.05} & \meanstd{.575}{.07}
 & \meanstd{.959}{.01} & \meanstd{.868}{.02} & \meanstd{.423}{.02} & \meanstd{.847}{.01} & \meanstd{.686}{.05} & \meanstd{.500}{.05} \\ \hline

\rowcolor{white}Omni-fMRI & $\mathbf{X}^V_g$ \rowstrut
 & \meanstd{.655}{.05} & \meanstd{.591}{.04} & \meanstd{.706}{.03} & \meanstd{.598}{.03} & \meanstd{.668}{.06} & \meanstd{\underline{.591}}{.07}
 & \meanstd{.958}{.01} & \meanstd{.859}{.02} & \meanstd{.392}{.01} & \meanstd{.862}{.01} & \meanstd{.676}{.05} & \meanstd{.514}{.02} \\ \hline

\rowcolor{white}CortexMAE & $\mathbf{X}^F_{cf}$ \rowstrut
 & \meanstd{.681}{.04} & \meanstd{\textbf{.618}}{.05} & \meanstd{.732}{.03} & \meanstd{\underline{.623}}{.03} & \meanstd{.645}{.06} & \meanstd{.555}{.06}
 & \meanstd{.928}{.02} & \meanstd{.811}{.02} & \meanstd{.558}{.05} & \meanstd{.720}{.06} & \meanstd{\underline{.653}}{.04} & \meanstd{.539}{.05} \\ \hline

\rowcolor{lprow}
\oursl & $\mathbf{X}_{g}^V, \mathbf{X}_{cf}^V$ \rowstrut
& \meanstd{\underline{.686}}{.05} & \meanstd{.604}{.05}
& \meanstd{\underline{.743}}{.03} & \meanstd{\textbf{.629}}{.04}
& \meanstd{\underline{.690}}{.04} & \meanstd{\underline{.591}}{.05}
& \meanstd{\textbf{.995}}{.00} & \meanstd{\textbf{.956}}{.03}
& \meanstd{\textbf{.320}}{.01} & \meanstd{\textbf{.917}}{.01}
& \meanstd{\textbf{.633}}{.06} & \meanstd{\textbf{.607}}{.05} \\

\rowcolor{lprow}
\oursl & $\mathbf{X}_{g}^F, \mathbf{X}_{cf}^F$ \rowstrut
& \meanstd{\textbf{.717}}{.03} & \meanstd{.607}{.05}
& \meanstd{\textbf{.747}}{.03} & \meanstd{.615}{.05}
& \meanstd{.679}{.05} & \meanstd{.541}{.04}
& \meanstd{\underline{.969}}{.00} & \meanstd{\underline{.884}}{.00}
& \meanstd{.476}{.02} & \meanstd{.828}{.01}
& \meanstd{.655}{.03} & \meanstd{\underline{.576}}{.04} \\
\hline
\end{tabular}
}
\label{tab:resting_partial_ft_results}
\end{table*}
To examine whether the trends in \Cref{tab:resting_ft_results} also hold under partial fine-tuning, we fine-tune only the final layer (one last block) of each foundation model together with its prediction head, freezing all remaining layers, except for Brain-DiT.
For Brain-DiT, we consider the last four blocks because the downstream representation aggregates hidden states from the 13th, 15th, and 16th blocks of the backbone, with the 13th block defining the earliest layer involved in downstream feature extraction.

As shown in \Cref{tab:resting_partial_ft_results}, partial fine-tuning
also fails to consistently outperform {\oursl} across tasks.

\subsection{Linear Probing Results with Matched Splits for Foundation Models}
\label{subsec:fixed_split}
\definecolor{lprow}{HTML}{FFF6CC}

\begin{table*}[!ht]
\caption{
Comparison of trait prediction performance. Each foundation model is evaluated both by linear probing on its frozen features and by full fine-tuning, while {\oursl} uses linear probing. Results are reported as the mean and standard deviation across data splits. \textbf{Bold} and \underline{underlined} values indicate the best and second-best performance for each metric, respectively. LP and FT denote linear probing and full fine-tuning, respectively.
}
\centering
\renewcommand{\arraystretch}{1.25}
\resizebox{\linewidth}{!}{%
\begin{tabular}{l|c|c|cc|cc|cc|cc|cc|cc}
\hline
\multirow{3}{*}{Method} & \multirow{3}{*}{Input} & \multirow{3}{*}{Eval}
& \multicolumn{2}{c|}{ADHD-200}
& \multicolumn{2}{c|}{ABIDE-II}
& \multicolumn{2}{c|}{ADNI}
& \multicolumn{6}{c}{HCP-A} \\ \cline{4-15}
 & & & \multicolumn{2}{c|}{Diagnosis}
 & \multicolumn{2}{c|}{Diagnosis}
 & \multicolumn{2}{c|}{Diagnosis}
 & \multicolumn{2}{c|}{Sex}
 & \multicolumn{2}{c|}{Age}
 & \multicolumn{2}{c}{Intelligence} \\
 & & & AUC $\uparrow$ & \multicolumn{1}{c|}{F1 $\uparrow$}
 & AUC $\uparrow$ & \multicolumn{1}{c|}{F1 $\uparrow$}
 & AUC $\uparrow$ & \multicolumn{1}{c|}{F1 $\uparrow$}
 & AUC $\uparrow$ & \multicolumn{1}{c|}{F1 $\uparrow$}
 & MAE $\downarrow$ & \multicolumn{1}{c|}{$\rho\uparrow$}
 & MAE $\downarrow$ & $\rho\uparrow$ \\ \hline

Brain-JEPA & $\mathbf{X}^P$ & \rowstrut LP
& \meanstd{.660}{.04} & \meanstd{.589}{.04}
& \meanstd{.728}{.03} & \meanstd{.623}{.04}
& \meanstd{.693}{.03} & \meanstd{.564}{.05}
& \meanstd{.892}{.03} & \meanstd{.785}{.04}
& \meanstd{.516}{.01} & \meanstd{.773}{.01}
& \meanstd{.715}{.05} & \meanstd{.407}{.05} \\
Brain-JEPA & $\mathbf{X}^P$ & \rowstrut FT
& \meanstd{.660}{.00} & \meanstd{.606}{.01}
& \meanstd{.720}{.05} & \meanstd{.622}{.04}
& \meanstd{.693}{.07} & \meanstd{.597}{.05}
& \meanstd{.917}{.00} & \meanstd{.829}{.00}
& \meanstd{.418}{.00} & \meanstd{.849}{.00}
& \meanstd{.681}{.01} & \meanstd{.487}{.01} \\ \hline

BH-F & $\mathbf{X}^P$ & \rowstrut LP
& \meanstd{.669}{.03} & \meanstd{\underline{.616}}{.02}
& \meanstd{.731}{.03} & \meanstd{.608}{.03}
& \meanstd{.601}{.05} & \meanstd{.432}{.07}
& \meanstd{.872}{.01} & \meanstd{.759}{.02}
& \meanstd{.434}{.00} & \meanstd{.840}{.00}
& \meanstd{.677}{.04} & \meanstd{.520}{.05} \\
BH-F & $\mathbf{X}^P$ & \rowstrut FT
& \meanstd{.631}{.05} & \meanstd{.573}{.06}
& \meanstd{.697}{.03} & \meanstd{.592}{.05}
& \meanstd{.601}{.06} & \meanstd{.452}{.12}
& \meanstd{.867}{.01} & \meanstd{.729}{.10}
& \meanstd{.435}{.02} & \meanstd{.838}{.02}
& \meanstd{.679}{.04} & \meanstd{.498}{.06} \\ \hline

Brain-DiT & $\mathbf{X}^P$ & \rowstrut LP
& \meanstd{.624}{.04} & \meanstd{.563}{.04}
& \meanstd{.675}{.03} & \meanstd{.547}{.03}
& \meanstd{.644}{.05} & \meanstd{.553}{.06}
& \meanstd{.925}{.02} & \meanstd{.828}{.03}
& \meanstd{.423}{.02} & \meanstd{.852}{.01}
& \meanstd{.675}{.03} & \meanstd{.514}{.05} \\
Brain-DiT & $\mathbf{X}^P$ & \rowstrut FT
& \meanstd{.648}{.05} & \meanstd{.568}{.04}
& \meanstd{.712}{.04} & \meanstd{.535}{.08}
& \meanstd{.659}{.05} & \meanstd{.484}{.14}
& \meanstd{.923}{.02} & \meanstd{.833}{.03}
& \meanstd{.386}{.01} & \meanstd{.877}{.00}
& \meanstd{.674}{.01} & \meanstd{.546}{.01} \\ \hline

SwiFT & $\mathbf{X}_g^V$ & \rowstrut LP
& \meanstd{.651}{.05} & \meanstd{.606}{.05}
& \meanstd{.705}{.03} & \meanstd{.603}{.04}
& \meanstd{\textbf{.708}}{.05} & \meanstd{.598}{.05}
& \meanstd{.913}{.02} & \meanstd{.801}{.03}
& \meanstd{.437}{.02} & \meanstd{.837}{.01}
& \meanstd{.683}{.03} & \meanstd{.485}{.03} \\
SwiFT & $\mathbf{X}_g^V$ & \rowstrut FT
& \meanstd{.664}{.05} & \meanstd{\textbf{.633}}{.05}
& \meanstd{.711}{.03} & \meanstd{.606}{.04}
& \meanstd{.688}{.07} & \meanstd{.582}{.06}
& \meanstd{.946}{.01} & \meanstd{.853}{.02}
& \meanstd{.350}{.03} & \meanstd{.897}{.02}
& \meanstd{.678}{.03} & \meanstd{.507}{.02} \\ \hline

NeuroSTORM & $\mathbf{X}_g^V$ & \rowstrut LP
& \meanstd{.645}{.05} & \meanstd{.599}{.05}
& \meanstd{.697}{.02} & \meanstd{.595}{.03}
& \meanstd{\underline{.701}}{.06} & \meanstd{\underline{.600}}{.05}
& \meanstd{.924}{.02} & \meanstd{.828}{.02}
& \meanstd{.460}{.02} & \meanstd{.822}{.01}
& \meanstd{.696}{.05} & \meanstd{.473}{.05} \\
NeuroSTORM & $\mathbf{X}_g^V$ & \rowstrut FT
& \meanstd{.658}{.04} & \meanstd{.572}{.07}
& \meanstd{.716}{.03} & \meanstd{.606}{.02}
& \meanstd{.673}{.04} & \meanstd{\textbf{.607}}{.02}
& \meanstd{.974}{.01} & \meanstd{.888}{.02}
& \meanstd{\underline{.343}}{.02} & \meanstd{\underline{.906}}{.01}
& \meanstd{.670}{.04} & \meanstd{.525}{.05} \\ \hline

Omni-fMRI & $\mathbf{X}_g^V$ & \rowstrut LP
& \meanstd{.657}{.04} & \meanstd{.609}{.03}
& \meanstd{.702}{.03} & \meanstd{.602}{.04}
& \meanstd{.674}{.04} & \meanstd{.588}{.04}
& \meanstd{.969}{.01} & \meanstd{.873}{.04}
& \meanstd{.388}{.02} & \meanstd{.875}{.01}
& \meanstd{\underline{.645}}{.04} & \meanstd{.552}{.03} \\
Omni-fMRI & $\mathbf{X}_g^V$ & \rowstrut FT
& \meanstd{.676}{.05} & \meanstd{.598}{.05}
& \meanstd{.712}{.03} & \meanstd{.586}{.03}
& \meanstd{.681}{.06} & \meanstd{.584}{.06}
& \meanstd{\underline{.975}}{.01} & \meanstd{\underline{.910}}{.01}
& \meanstd{.351}{.01} & \meanstd{.894}{.01}
& \meanstd{.657}{.04} & \meanstd{.546}{.05} \\ \hline

CortexMAE & $\mathbf{X}_{cf}^F$ & \rowstrut LP
& \meanstd{.677}{.03} & \meanstd{.613}{.04}
& \meanstd{.730}{.02} & \meanstd{.612}{.03}
& \meanstd{.652}{.05} & \meanstd{.569}{.06}
& \meanstd{.947}{.01} & \meanstd{.858}{.02}
& \meanstd{.536}{.02} & \meanstd{.739}{.02}
& \meanstd{.655}{.04} & \meanstd{.553}{.05} \\
CortexMAE & $\mathbf{X}_{cf}^F$ & \rowstrut FT
& \meanstd{\underline{.690}}{.05} & \meanstd{.605}{.06}
& \meanstd{.735}{.03} & \meanstd{\textbf{.639}}{.04}
& \meanstd{.638}{.05} & \meanstd{.540}{.06}
& \meanstd{.944}{.02} & \meanstd{.859}{.02}
& \meanstd{.508}{.02} & \meanstd{.774}{.02}
& \meanstd{.661}{.05} & \meanstd{.539}{.06} \\ \hline

\rowcolor{lprow}
\oursl & $\mathbf{X}_{g}^V, \mathbf{X}_{cf}^V$ & \rowstrut LP
& \meanstd{.686}{.05} & \meanstd{.604}{.05}
& \meanstd{\underline{.743}}{.03} & \meanstd{\underline{.629}}{.04}
& \meanstd{.690}{.04} & \meanstd{.591}{.05}
& \meanstd{\textbf{.995}}{.00} & \meanstd{\textbf{.956}}{.03}
& \meanstd{\textbf{.320}}{.01} & \meanstd{\textbf{.917}}{.01}
& \meanstd{\textbf{.633}}{.06} & \meanstd{\textbf{.607}}{.05} \\

\rowcolor{lprow}
\oursl & $\mathbf{X}_{g}^F, \mathbf{X}_{cf}^F$ & \rowstrut LP
& \meanstd{\textbf{.717}}{.03} & \meanstd{.607}{.05}
& \meanstd{\textbf{.747}}{.03} & \meanstd{.615}{.05}
& \meanstd{.679}{.05} & \meanstd{.541}{.04}
& \meanstd{.969}{.00} & \meanstd{.884}{.00}
& \meanstd{.476}{.02} & \meanstd{.828}{.01}
& \meanstd{.655}{.03} & \meanstd{\underline{.576}}{.04} \\
\hline
\end{tabular}}
\label{tab:fixed_split_ft_lb}
\vspace{-0.15in}
\end{table*}
The linear-probing comparison in \Cref{fig:lp_results} uses 500
randomly sampled splits, whereas the fine-tuning results in
\Cref{tab:resting_ft_results} use the 12 or four splits described in
\Cref{subsec:experimental_settings}, so the two are not directly
comparable.
To compare the two adaptation strategies on identical partitions, we
additionally linear-probe every foundation model on the same splits
used for fine-tuning.

From \Cref{tab:fixed_split_ft_lb}, it can be observed that fine-tuning is not uniformly better: linear probing is ahead on ADNI for SwiFT and NeuroSTORM, on ABIDE-II and ADHD-200 for BrainHarmonix-F, and on intelligence regression for Omni-fMRI and CortexMAE, while fine-tuning helps most consistently on the HCP-A sex and age tasks.
Adapting the entire pre-trained encoder therefore does not reliably
extract more from these representations than a linear readout does,
and neither protocol brings the foundation models past {\oursl} on most
tasks.

\subsection{Full Results of \texorpdfstring{\Cref{fig:lp_results}}{lp results} and \texorpdfstring{\Cref{fig:hbn_results}}{hbn results}}

We present full results of \Cref{fig:lp_results} and \Cref{fig:hbn_results} in \Cref{tab:lp_full_results} and \Cref{tab:full_hbn_movie_results}, respectively.

\definecolor{lprow}{HTML}{FFF6CC}

\begin{table*}[!ht]
\caption{
Full results of \Cref{fig:lp_results}. \textbf{Bold} and \underline{underlined} values indicate the best and second-best performance for each metric, respectively.
}
\vspace{-0.1in}
\centering
\renewcommand{\arraystretch}{1.25}
\resizebox{\linewidth}{!}{%
\begin{tabular}{l|c|cc|cc|cc|cc|cc|cc}
\hline
\multirow{3}{*}{Method} & \multirow{3}{*}{Input} & \multicolumn{2}{c|}{ADHD-200} & \multicolumn{2}{c|}{ABIDE-II} & \multicolumn{2}{c|}{ADNI} & \multicolumn{6}{c}{HCP-A} \\ \cline{3-14}
 & & \multicolumn{2}{c|}{Diagnosis} & \multicolumn{2}{c|}{Diagnosis} & \multicolumn{2}{c|}{Diagnosis} & \multicolumn{2}{c|}{Sex} & \multicolumn{2}{c|}{Age} & \multicolumn{2}{c}{Intelligence} \\
 & & AUC $\uparrow$ & \multicolumn{1}{c|}{F1 $\uparrow$} & AUC $\uparrow$ & \multicolumn{1}{c|}{F1 $\uparrow$} & AUC $\uparrow$ & \multicolumn{1}{c|}{F1 $\uparrow$} & AUC $\uparrow$ & \multicolumn{1}{c|}{F1 $\uparrow$} & MAE $\downarrow$ & \multicolumn{1}{c|}{$\rho\uparrow$} & MAE $\downarrow$ & $\rho\uparrow$ \\ \hline

BrainMASS & $\mathbf{X}^P$ \rowstrut
& \meanstd{.638}{.06} & \meanstd{.552}{.07} & \meanstd{.695}{.04} & \meanstd{.596}{.05} & \meanstd{.637}{.07} & \meanstd{.533}{.07} & \meanstd{.823}{.03} & \meanstd{.699}{.04} & \meanstd{.635}{.03} & \meanstd{.628}{.04} & \meanstd{.679}{.04} & \meanstd{.493}{.06} \\
Brain-JEPA & $\mathbf{X}^P$ \rowstrut
& \meanstd{.655}{.04} & \meanstd{.589}{.04} & \meanstd{.720}{.03} & \meanstd{.619}{.04} & \meanstd{\textbf{.684}}{.06} & \meanstd{.561}{.07} & \meanstd{.874}{.03} & \meanstd{.763}{.04} & \meanstd{.502}{.03} & \meanstd{.784}{.02} & \meanstd{.712}{.04} & \meanstd{.418}{.06} \\
BH-F & $\mathbf{X}^P$ \rowstrut
& \meanstd{.662}{.03} & \meanstd{\textbf{.631}}{.02} & \meanstd{.740}{.03} & \meanstd{.621}{.04} & \meanstd{.619}{.07} & \meanstd{.453}{.09} & \meanstd{.883}{.03} & \meanstd{.770}{.04} & \meanstd{.420}{.02} & \meanstd{.854}{.02} & \meanstd{.682}{.04} & \meanstd{.507}{.05} \\
Brain-DiT & $\mathbf{X}^P$ \rowstrut
& \meanstd{.633}{.04} & \meanstd{.586}{.03} & \meanstd{.698}{.03} & \meanstd{.577}{.04} & \meanstd{.651}{.06} & \meanstd{.542}{.07} & \meanstd{.925}{.02} & \meanstd{.823}{.03} & \meanstd{.419}{.02} & \meanstd{.855}{.02} & \meanstd{.680}{.04} & \meanstd{.501}{.06} \\
\hline

SwiFT & $\mathbf{X}_g^V$ \rowstrut
& \meanstd{.649}{.04} & \meanstd{.600}{.04} & \meanstd{.703}{.03} & \meanstd{.600}{.04} & \meanstd{.678}{.06} & \meanstd{\textbf{.577}}{.07} & \meanstd{.905}{.02} & \meanstd{.797}{.03} & \meanstd{.424}{.02} & \meanstd{.852}{.02} & \meanstd{.679}{.04} & \meanstd{.491}{.05} \\
NeuroSTORM & $\mathbf{X}_g^V$ \rowstrut
& \meanstd{.630}{.04} & \meanstd{.587}{.04} & \meanstd{.712}{.03} & \meanstd{.602}{.04} & \meanstd{.668}{.06} & \meanstd{.562}{.07} & \meanstd{.918}{.02} & \meanstd{.805}{.03} & \meanstd{.445}{.02} & \meanstd{.837}{.02} & \meanstd{.687}{.04} & \meanstd{.490}{.05} \\
Omni-fMRI & $\mathbf{X}_g^V$ \rowstrut
& \meanstd{.654}{.04} & \meanstd{.598}{.05} & \meanstd{.723}{.03} & \meanstd{.612}{.04} & \meanstd{.654}{.06} & \meanstd{.551}{.07} & \meanstd{.960}{.01} & \meanstd{.875}{.03} & \meanstd{.386}{.02} & \meanstd{.877}{.01} & \meanstd{.655}{.04} & \meanstd{.545}{.05} \\
\hline

CortexMAE & $\mathbf{X}_{cf}^F$ \rowstrut
& \meanstd{.670}{.04} & \meanstd{\underline{.618}}{.05} & \meanstd{.727}{.04} & \meanstd{.595}{.05} & \meanstd{.654}{.06} & \meanstd{.568}{.06} & \meanstd{.949}{.02} & \meanstd{.856}{.03} & \meanstd{.529}{.03} & \meanstd{.754}{.03} & \meanstd{.665}{.04} & \meanstd{.540}{.05} \\
\hline


\oursl & $\mathbf{X}_{g}^V$ \rowstrut
& \meanstd{.678}{.04} & \meanstd{.603}{.05} & \meanstd{.744}{.03} & \meanstd{\textbf{.630}}{.05} & \meanstd{.680}{.06} & \meanstd{\underline{.571}}{.07} & \meanstd{\underline{.988}}{.01} & \meanstd{\underline{.948}}{.02} & \meanstd{\underline{.328}}{.02} & \meanstd{\underline{.914}}{.01} & \meanstd{.644}{.04} & \meanstd{.582}{.04} \\

\oursl & $\mathbf{X}_{cf}^V$ \rowstrut
& \meanstd{.676}{.04} & \meanstd{.612}{.05} & \meanstd{.746}{.04} & \meanstd{.625}{.05} & \meanstd{.676}{.06} & \meanstd{.571}{.07} & \meanstd{.985}{.01} & \meanstd{.931}{.02} & \meanstd{.348}{.02} & \meanstd{.900}{.01} & \meanstd{\underline{.616}}{.04} & \meanstd{\underline{.616}}{.04} \\

\oursl & $\mathbf{X}_{g}^V, \mathbf{X}_{cf}^V$ \rowstrut
& \meanstd{.682}{.04} & \meanstd{.609}{.05} & \meanstd{.749}{.03} & \meanstd{\underline{.628}}{.05} & \meanstd{\underline{.681}}{.06} & \meanstd{.563}{.07} & \meanstd{\textbf{.992}}{.01} & \meanstd{\textbf{.961}}{.02} & \meanstd{\textbf{.312}}{.02} & \meanstd{\textbf{.922}}{.01} & \meanstd{\textbf{.616}}{.04} & \meanstd{\textbf{.619}}{.04} \\

\oursl & $\mathbf{X}_{g}^F$ \rowstrut
& \meanstd{.687}{.04} & \meanstd{.617}{.05} & \meanstd{.718}{.03} & \meanstd{.618}{.04} & \meanstd{.666}{.06} & \meanstd{.551}{.07} & \meanstd{.935}{.02} & \meanstd{.833}{.03} & \meanstd{.509}{.03} & \meanstd{.780}{.02} & \meanstd{.695}{.04} & \meanstd{.491}{.05} \\

\oursl & $\mathbf{X}_{cf}^F$ \rowstrut
& \meanstd{\underline{.704}}{.04} & \meanstd{.593}{.07} & \meanstd{\underline{.754}}{.03} & \meanstd{.618}{.05} & \meanstd{.674}{.06} & \meanstd{.560}{.08} & \meanstd{.945}{.02} & \meanstd{.855}{.03} & \meanstd{.491}{.03} & \meanstd{.792}{.02} & \meanstd{.648}{.04} & \meanstd{.565}{.05} \\

\oursl & $\mathbf{X}_{g}^F, \mathbf{X}_{cf}^F$ \rowstrut
& \meanstd{\textbf{.708}}{.04} & \meanstd{.612}{.06} & \meanstd{\textbf{.755}}{.03} & \meanstd{.617}{.04} & \meanstd{.672}{.06} & \meanstd{.555}{.07} & \meanstd{.972}{.01} & \meanstd{.896}{.03} & \meanstd{.448}{.02} & \meanstd{.844}{.02} & \meanstd{.641}{.04} & \meanstd{.589}{.05} \\
\hline
\end{tabular}}
\label{tab:lp_full_results}
\vspace{-0.2in}
\end{table*}

\definecolor{lprow}{HTML}{FFF6CC}
\begin{table*}[!ht]
\caption{
Full results of \Cref{fig:hbn_results}. LP rows are highlighted in yellow; FT and TFS denote fine-tuning and training from scratch, respectively. \textbf{Bold} and \underline{underlined} values indicate the best and second-best distinct performance for each metric, respectively.
}
\vspace{-0.1in}
\centering
\renewcommand{\arraystretch}{1.25}
\resizebox{0.5\linewidth}{!}{%
\begin{tabular}{l|c|c|cc}
\hline
\multirow{2}{*}{Method} & \multirow{2}{*}{Eval.} & \multirow{2}{*}{Input} & \multicolumn{2}{c}{HBN-Movie} \\ \cline{4-5}
 & & & AUC $\uparrow$ & F1 $\uparrow$ \\ \hline

\rowcolor{lprow}
BrainMASS & LP & $\mathbf{X}^{P}$ \rowstrut
& \meanstd{.895}{.02} & \meanstd{.793}{.02} \\
\rowcolor{lprow}
Brain-JEPA & LP & $\mathbf{X}^{P}$ \rowstrut
& \meanstd{.631}{.02} & \meanstd{.610}{.04} \\
Brain-JEPA & FT & $\mathbf{X}^{P}$ \rowstrut
& \meanstd{.716}{.10} & \meanstd{.656}{.08} \\
\rowcolor{lprow}
BH-F & LP & $\mathbf{X}^{P}$ \rowstrut
& \meanstd{.724}{.04} & \meanstd{.647}{.05} \\
BH-F & FT & $\mathbf{X}^{P}$ \rowstrut
& \meanstd{.707}{.18} & \meanstd{.673}{.16} \\
\rowcolor{lprow}
Brain-DiT & LP & $\mathbf{X}^{P}$ \rowstrut
& \meanstd{.739}{.03} & \meanstd{.675}{.04} \\
Brain-DiT & FT & $\mathbf{X}^{P}$ \rowstrut
& \meanstd{.897}{.13} & \meanstd{.691}{.13} \\
\hline

\rowcolor{lprow}
SwiFT & LP & $\mathbf{X}_{g}^V$ \rowstrut
& \meanstd{.630}{.04} & \meanstd{.599}{.02} \\
\rowcolor{lprow}
SwiFT & LP & $\mathbf{X}_{cf}^V$ \rowstrut
& \meanstd{.708}{.02} & \meanstd{.663}{.00} \\
SwiFT & FT & $\mathbf{X}_{g}^V$ \rowstrut
& \meanstd{.619}{.05} & \meanstd{.590}{.06} \\
SwiFT & FT & $\mathbf{X}_{cf}^V$ \rowstrut
& \meanstd{.633}{.06} & \meanstd{.602}{.04} \\
\rowcolor{lprow}
NeuroSTORM & LP & $\mathbf{X}_{g}^V$ \rowstrut
& \meanstd{.632}{.04} & \meanstd{.597}{.04} \\
\rowcolor{lprow}
NeuroSTORM & LP & $\mathbf{X}_{cf}^V$ \rowstrut
& \meanstd{.739}{.02} & \meanstd{.692}{.05} \\
NeuroSTORM & FT & $\mathbf{X}_{g}^V$ \rowstrut
& \meanstd{.620}{.06} & \meanstd{.538}{.08} \\
NeuroSTORM & FT & $\mathbf{X}_{cf}^V$ \rowstrut
& \meanstd{.690}{.09} & \meanstd{.659}{.08} \\
\rowcolor{lprow}
Omni-fMRI & LP & $\mathbf{X}_{g}^V$ \rowstrut
& \meanstd{.615}{.03} & \meanstd{.591}{.03} \\
\rowcolor{lprow}
Omni-fMRI & LP & $\mathbf{X}_{cf}^V$ \rowstrut
& \meanstd{.783}{.03} & \meanstd{.713}{.02} \\
Omni-fMRI & FT & $\mathbf{X}_{g}^V$ \rowstrut
& \meanstd{.596}{.06} & \meanstd{.561}{.10} \\
Omni-fMRI & FT & $\mathbf{X}_{cf}^V$ \rowstrut
& \meanstd{.928}{.03} & \meanstd{.838}{.02} \\
\hline

\rowcolor{lprow}
CortexMAE & LP & $\mathbf{X}_{cf}^F$ \rowstrut
& \meanstd{.939}{.02} & \meanstd{.852}{.06} \\
CortexMAE & FT & $\mathbf{X}_{cf}^F$ \rowstrut
& \meanstd{\underline{.997}}{.00} & \meanstd{.959}{.01} \\
\hline

TABLeT & TFS & $\mathbf{X}_{g}^V$ \rowstrut
& \meanstd{.609}{.05} & \meanstd{.499}{.19} \\
TABLeT & TFS & $\mathbf{X}_{cf}^V$ \rowstrut
& \meanstd{.990}{.01} & \meanstd{.948}{.03} \\
\hline

\rowcolor{lprow}
FReD-Trait & LP & $\mathbf{X}_{g}^V, \mathbf{X}_{cf}^V$ \rowstrut
& \meanstd{.849}{.03} & \meanstd{.780}{.03} \\
\rowcolor{lprow}
FReD-Trait & LP & $\mathbf{X}_{g}^V$ \rowstrut
& \meanstd{.672}{.03} & \meanstd{.623}{.04} \\
\rowcolor{lprow}
FReD-Trait & LP & $\mathbf{X}_{cf}^V$ \rowstrut
& \meanstd{.852}{.03} & \meanstd{.784}{.03} \\
\rowcolor{lprow}
FReD-Trait & LP & $\mathbf{X}_{g}^F, \mathbf{X}_{cf}^F$ \rowstrut
& \meanstd{.871}{.02} & \meanstd{.808}{.01} \\
\rowcolor{lprow}
FReD-Trait & LP & $\mathbf{X}_{g}^F$ \rowstrut
& \meanstd{.629}{.03} & \meanstd{.580}{.04} \\
\rowcolor{lprow}
FReD-Trait & LP & $\mathbf{X}_{cf}^F$ \rowstrut
& \meanstd{.871}{.02} & \meanstd{.802}{.01} \\
\hline

FReD-State & TFS & $\mathbf{X}_{cf}^V$ \rowstrut
& \meanstd{\textbf{.998}}{.00} & \meanstd{\textbf{.973}}{.01} \\
FReD-State & TFS & $\mathbf{X}_{cf}^F$ \rowstrut
& \meanstd{\underline{.997}}{.00} & \meanstd{\underline{.971}}{.01} \\
\hline
\end{tabular}}
\label{tab:full_hbn_movie_results}
\vspace{-0.2in}
\end{table*}

\section{Architectural Comparison between TABLeT and \ourst}
\label{sec:tablet_arch}

\Cref{tab:tablet_v1_vs_v2_arch} lists the differences between TABLeT and {\ourst} with $\mathbf{X}^{V}$ in full.
Both models take the same DCAE latent tensor and the same downstream loss, so every difference below concerns how those latents are turned into a sequence and how that sequence is modeled.
The ablation in \Cref{tab:tablet_v1_vs_v2} isolates the contribution of tokenization, while the remaining architectural and regularization changes are evaluated jointly.

\paragraph{Tokenization.}
The two models differ in whether the slice axis is aggregated before or
after the temporal Transformer.
TABLeT keeps three slice blocks and the $3\times3$ latent grid as
separate tokens, giving 27 tokens per frame and a sequence of length
$27T+1$, so the Transformer must resolve spatial and temporal structure
jointly.
{\ourst} instead averages within slice groups and flattens all axes,
groups, channels, and spatial positions into a single vector per frame, which a linear layer projects to the
model width.
The sequence is therefore $T+1$ tokens long, and spatial mixing happens
in the projection rather than in self-attention.
This reduces the number of self-attention token pairs significantly and lets the Transformer allocate its capacity to temporal structure alone.

\paragraph{Transformer backbone.}
{\ourst} uses a third of the depth, four blocks instead of twelve, and
adopts several standard modernizations: RMSNorm in place of LayerNorm, SwiGLU \citep{swiglu} feed-forward blocks in place of a ReLU MLP, and Q/K normalization \citep{qknorm} with soft-capped attention logits \citep{soft_cap}. Model and feed-forward widths are unchanged, and both models use a
learned \texttt{[CLS]} token with rotary position embeddings \citep{rope}.

\paragraph{Regularization.}
Since a frame is now a single token, dropping frames during training is a natural regularizer, and we drop frame tokens with $p=0.5$ while never dropping the \texttt{[CLS]} token.
We additionally use stochastic depth \citep{drop_path} with a drop-path probability increasing linearly from 0 to 0.1 across the four blocks.

Together these changes reduce the trainable parameter count from 129.5M to 78.2M, a 39.6\% reduction, while improving accuracy on both HCP-Task and NSD (\Cref{tab:tablet_v1_vs_v2}).

\begin{table*}[t]
  \centering
  \caption{
    Architectural comparison between TABLeT and {\ourst} with
    $\mathbf{X}^V$. Both models receive the same DCAE latent tensor and
    use the same downstream loss; $T$ denotes the number of input fMRI
    frames.
  }
  \label{tab:tablet_v1_vs_v2_arch}
  \small
  \setlength{\tabcolsep}{5pt}
  \renewcommand{\arraystretch}{1.15}
  \begin{tabular}{
    @{}
    >{\raggedright\arraybackslash}p{0.21\textwidth}
    >{\raggedright\arraybackslash}p{0.36\textwidth}
    >{\raggedright\arraybackslash}p{0.36\textwidth}
    @{}
  }
    \toprule
    \textbf{Component}
      & \textbf{TABLeT}
      & \textbf{{\ourst} with $\mathbf{X}^V$} \\
    \midrule

    \multicolumn{3}{@{}l}{\textit{Input representation and tokenization}} \\
    \addlinespace[2pt]
    DCAE latent input
      & $\mathbf{x}\in\mathbb{R}^{T\times3\times96\times32\times3\times3}$
      & Same input tensor \\

    Intra-frame aggregation
      & No slice averaging. Three contiguous 32-slice blocks are formed per
        axis, and corresponding features from the three anatomical axes are
        concatenated at each latent-grid position.
      & Contiguous slice-group averaging with $G=24$ groups per axis
        ($96\rightarrow24$), followed by concatenation of all axes,
        groups, channels, and spatial locations into one frame vector. \\

    Tokens per frame
      & $3$ slice blocks $\times\,3\times3$ spatial positions $=27$
      & $1$ whole-frame token \\

    Pre-projection feature width
      & $3{,}072$
      & $20{,}736$ \\

    Frame projection
      & $\operatorname{LN}(3072)\rightarrow
        \operatorname{Linear}(3072,896) \rightarrow \operatorname{LN}(896)$
      & $\operatorname{LN}(20736)\rightarrow
        \operatorname{Linear}(20736,896)\rightarrow
        \operatorname{LN}(896)$ \\

    Sequence length with \textsc{cls}
      & $27T+1$
      & $T+1$ \\

    \midrule
    \multicolumn{3}{@{}l}{\textit{Temporal Transformer backbone}} \\
    \addlinespace[2pt]
    Transformer depth
      & $12$ blocks
      & $4$ blocks \\

    Model and FFN widths
      & $d_{\mathrm{model}}=896$, $d_{\mathrm{ff}}=4864$
      & $d_{\mathrm{model}}=896$, $d_{\mathrm{ff}}=4864$ \\

    Grouped-query attention
      & $14$ query heads, $2$ key/value heads; head dimension $64$
      & $16$ query heads, $2$ key/value heads; head dimension $56$ \\

    Attention formulation
      & Scaled dot-product GQA
      & Q/K RMS normalization and soft-capped attention logits,
        $10\tanh(\ell/10)$, before softmax \\

    Positional encoding
      & One-dimensional RoPE
      & One-dimensional RoPE \\

    Transformer normalization
      & Pre-norm LayerNorm
      & Pre-norm RMSNorm \\

    Feed-forward network
      & Vanilla two-layer MLP with ReLU:
        $896\rightarrow4864\rightarrow896$
      & SwiGLU with separate gate and value projections and a SiLU gate:
        $896\rightarrow4864\rightarrow896$ \\

    Readout
      & Learned \texttt{[CLS]} token followed by a linear task head
      & Learned \texttt{[CLS]} token followed by a linear task head \\

    \midrule
    \multicolumn{3}{@{}l}{\textit{Regularization and computational scale}} \\
    \addlinespace[2pt]
    Frame dropout
      & None
      & Training-only frame-token dropout with $p=0.5$ \\

    Stochastic depth
      & None
      & Drop-path probability increases linearly from $0$ to $0.1$ across
        the four blocks \\

    Trainable parameters
      & $129{,}533{,}333$ ($129.53$M)
      & $78{,}225{,}493$ ($78.23$M; $39.6\%$ fewer) \\
    \bottomrule
  \end{tabular}
\end{table*}

\end{document}